\documentclass[11pt]{article}

\usepackage[preprint]{acl}

\usepackage{times}
\usepackage{latexsym}

\usepackage[T1]{fontenc}
\usepackage[utf8]{inputenc}

\usepackage{microtype}
\usepackage{xcolor}

\usepackage{inconsolata}

\usepackage{graphicx}
\usepackage{hyperref}
\usepackage{url}
\usepackage{times}
\usepackage{latexsym}
\usepackage{multirow}
\usepackage{booktabs}
\usepackage{amsfonts}
\usepackage{amsmath}
{

}

\usepackage{booktabs,array,xcolor}
\usepackage{enumitem}
\usepackage{subcaption}
\usepackage{wrapfig}
\usepackage{booktabs,subcaption}
\usepackage{amssymb} 

\definecolor{grey}{gray}{0.85}
\definecolor{rowgray}{gray}{0.95}
\definecolor{headergray}{gray}{0.85}

\newcommand*\colourcheck[1]{%
  \expandafter\newcommand\csname #1check\endcsname{\textcolor{#1}{\ding{52}}}%
}
\colourcheck{blue}
\colourcheck{green}
\colourcheck{red}
\usepackage{pifont}
\newcommand{\xmark}{\ding{55}}% 

\newcommand{\best}[1]{\textbf{#1}}
\newcommand{\ds}[1]{\textsc{#1}}
\usepackage[most]{tcolorbox}
\definecolor{grey}{gray}{0.85}

\usepackage{makecell}

\title{Monitorable Chart Reasoning Agents via Verifiable Process Rewards}

\author{
 \textbf{Sanchit Sinha\textsuperscript{1}},
 \textbf{Oana Frunza\textsuperscript{2}},
 \textbf{Kashif Rasul\textsuperscript{2}},
 \textbf{Aidong Zhang\textsuperscript{1}},
\\
 \textsuperscript{1}University of Virginia, Charlottesville, VA, USA,
 \textsuperscript{2}Morgan Stanley, USA
}

\begin{document}
\maketitle
\begin{abstract}
Chart reasoning agents are increasingly used to extract actionable insights in critical domains, achieving state-of-the-art performance on multiple benchmarks. Yet, high benchmark accuracy alone is insufficient for deployment, where stakeholders must be able to audit and verify how a model reaches its answer. Existing LVLM-based chart agents produce either answer-only predictions or free-form rationales that are hard to verify, obscuring whether an error arose from misreading the chart, extracting a wrong value, or miscomputing. We propose \textbf{Chart-RVR}, a reinforcement learning framework for training \textit{monitorable chart agents} with verifiable process rewards. Chart-RVR decomposes chart reasoning into three auditable blocks: \textit{Structure}, identifying the chart type; \textit{Evidence}, reconstructing the underlying data table in JSON; and \textit{Derivation}, exposing the stepwise trace that computes the answer. Across six in-domain and out-of-domain benchmarks, Chart-RVR attains state-of-the-art accuracy among comparable-sized LVLMs. Beyond accuracy, we assess monitorability using a triangulated protocol that combines ground-truth surrogate metrics, an oracle information-gain measure, and an LLM-as-auditor scoring \textit{Process Verifiability} and \textit{Evidence Localization}, showing that Chart-RVR yields rationales that are markedly more verifiable and evidence-grounded than those from CoT prompting, SFT, and existing chart-specific baselines. The code can be found at \url{https://github.com/sanchit97/chartrl}.
\end{abstract}
\section{Introduction}
\begin{figure*}[t]
    \centering
    \includegraphics[width=0.85\textwidth]{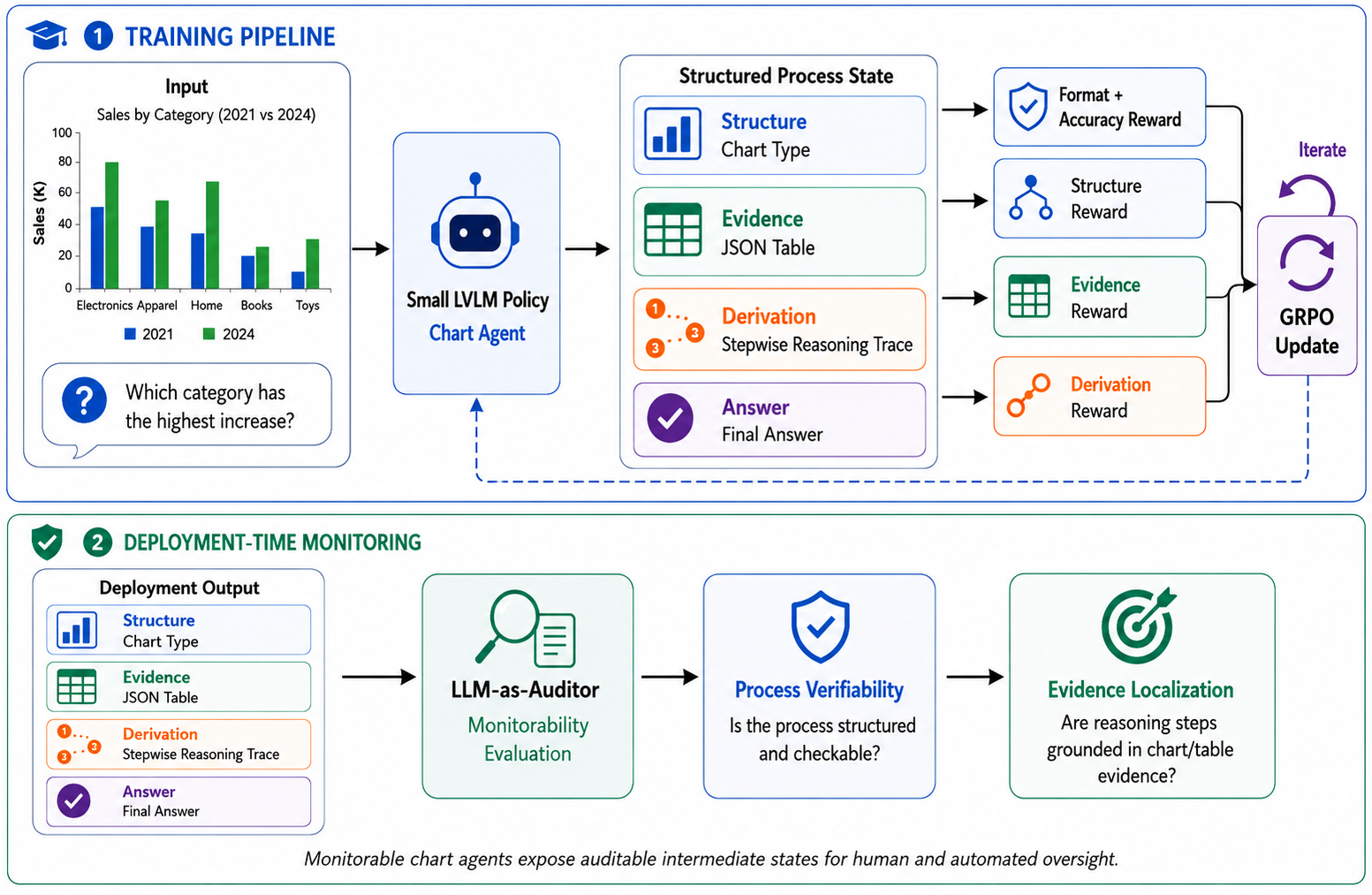}
    \caption{Overview of Chart-RVR. Top (training): a small LVLM policy emits a structured process state — chart type (Structure), reconstructed JSON table (Evidence), and a stepwise trace (Derivation) — and each block earns a verifiable reward that drives the GRPO update. Bottom (deployment): the same exposed blocks are scored by an LLM-as-auditor along Process Verifiability and Evidence Localization, so failures can be localized.}
    \label{fig:motivation}
    % \vspace{-10pt}
\end{figure*}
% % charts matter; chart reasoning is entangled; LVLMs do well on benchmarks
Charts are a cornerstone of visual communication across finance, healthcare, public policy, and beyond, where experts and non-experts alike rely on them to allocate resources and drive strategic decisions. Automating their interpretation is therefore a high-value problem. Unlike natural images, typically described by high-level semantics (e.g., ``a dog on a table''), charts encode information through \textit{precise} spatial and numerically aligned relationships. This makes chart reasoning inherently \textit{entangled}: structured data is tightly interwoven with visual design choices, and any chart-reasoning model must disentangle the two during decision making. Large Vision-Language Models (LVLMs), pre-trained on billions of image-text pairs, have shown strong performance on general visual question answering, including chart reasoning.

% % benchmark accuracy is insufficient for deployment; monitorability
However, high benchmark accuracy alone is insufficient for deploying chart agents in high-stakes settings. Stakeholders in sensitive domains like health and finance demand correct answers and also accurate \textit{intermediate evidence} and transparency into how it was reached. This entails understanding which \textit{chart structure} was recognized, what \textit{entities} were extracted, and how those entities were \textit{combined}. We refer to this property as \textbf{monitorability} in chart reasoning, an emerging area of deployable agentic research \citep{guan2025monitoring}. For chart agents, monitorability captures the extent to which intermediate reasoning is structured, verifiable, localizable, and understandable to end-users.

% % two categories of existing work and their weaknesses
The majority of existing chart-reasoning agents fall into two categories. The first uses off-the-shelf pre-trained LVLMs inside a systematic harness. While effective on standard benchmarks, recent studies \cite{islam2024large} expose two systematic weaknesses: even when these agents answer in-domain (ID) questions correctly, eliciting rationales via chain-of-thought (CoT) prompting often fails to improve accuracy and can even \textit{harm} it \citep{zhang2024improve,turpin2023language}, yielding incoherent or hallucinated traces. This brittleness is most acute for smaller LVLMs, which are attractive for efficient deployment but are the most unreliable at faithful multi-step reasoning. The second category fine-tunes agents on chart-specific data such as ChartGemma \citep{masry2024chartgemma} and BigCharts-R1 \citep{masrybigcharts}, which first curate gold rationales and train via Supervised Finetuning (SFT) or RL-based approaches like GRPO. Even still, their rationales remain only \textit{partially controllable}, and the rationale itself is never an explicit verification target - making it unsuitable for effective deployment.

% % free-form rationales are not enough; SFT imitates, answer-only GRPO is opaque
As a consequence, fine-tuning chart agents on the final answer is not enough. Even if the answer is correct, the rationale may read fluently but misclassify the chart type, read a hallucinated value, select the wrong evidence, or miscompute the steps taken to get to the answer. SFT on reasoning traces \citep{masry2024chartgemma,carbune2024chart,zhang2024tinychart} imitates demonstration tokens, but token-level imitation does not guarantee verifiability, as a model can reproduce dataset-specific explanation styles while still incorrectly reading intermediate values. Current systems utilizing Reinforcement Fine-Tuning, like GRPO-style \citep{guo2025deepseek,shao2024deepseekmath} outcome optimization, have shown to achieve high benchmark scores, but are even less monitorable as the rollouts generated are free form, without any constraints. Hence, systems like \citet{masrybigcharts} which use \textit{answer-level} GRPO remain insufficient for monitorability as it offers no signal about whether perception, extraction, evidence selection, or arithmetic failed.

% % the 3-block decomposition: Structure, Evidence, Derivation
To mitigate this problem, we posit that monitorable chart agents require decomposing the reasoning process into three verifiable blocks: \textit{Structure}, \textit{Evidence}, and \textit{Derivation}. The \textit{Structure} block identifies the chart type, trivially verified by visual inspection, confirming whether values should be read from bars, lines, or sectors. Subsequently, the \textit{Evidence} block reconstructs the underlying data table, exposing the precise values and labels used downstream and making hallucinated, missing, or inconsistent entries easy to detect. Finally, the \textit{Derivation} block captures the algorithmic trace that composes this evidence into the answer via localization, arithmetic, averaging, etc., i.e., the \textbf{process} to compute the final answer. Together, these blocks \textit{convert} unconstrained free-form reasoning into a structured process state that can be verified, localized, and monitored at deployment. A high-level schematic is shown in  Figure~\ref{fig:motivation}.

% % we propose Chart-RVR + rewards + delta % % empirical overview: no accuracy tax, controlled methodology win, OOD, cross-arch, triangulated monitorability
Building on this premise, we propose \textbf{Chart-RVR}, a reinforcement learning framework for training \textit{monitorable} chart agents with verifiable structure and process rewards. Chart-RVR pairs GRPO with chart-specific rewards that elicit a controlled CoT rationale comprising - (i) the chart type, (ii) underlying table reconstruction, (iii) a stepwise evidence-gathering and computation (derivation) trace, and (iv) the final answer. Unlike prior methods that curate gold rationales but reward only the final answer, Chart-RVR makes \textit{each block of the process an explicit, checkable reward target}, aligning the model with intermediate states that stakeholders can inspect and audit. Most CoT and curated-rationale baselines fall below even non-explanatory direct prompting, whereas Chart-RVR exposes a fully verifiable process while exceeding it across six benchmarks, with the largest margins under distribution shift. Beyond accuracy, we also propose quantifiable monitorability with a triangulated protocol: ground-truth \textit{surrogate metrics} (chart-type accuracy and table reconstruction), an \textit{oracle information-gain} measure quantifying how much a rationale raises a held-out LVLM's certainty in the correct answer, and an \textit{LLM-as-auditor} scoring \textit{Process Verifiability} and \textit{Evidence Localization}. Our contributions are as follows:
\begin{itemize}[leftmargin=*, parsep=0pt, itemsep=0pt, topsep=0pt]
    \item We formulate \textit{monitorable chart reasoning} as a deployment-oriented task in which models must expose structured intermediate states that support verification and evidence localization.

    \item We propose \textbf{Chart-RVR}, a GRPO-based framework that trains small LVLMs with verifiable process rewards over chart-type prediction, JSON table reconstruction, and algorithmic process conformity, making the full reasoning process, not just the final answer, a checkable target.

    \item We show that Chart-RVR attains state-of-the-art accuracy among comparable LVLMs across six benchmarks \textit{without} the explainability-accuracy tradeoff, with the strongest gains on Out of Domain (OOD) data with a comprehensive controlled reward-only comparison.

    \item We introduce a triangulated monitorability evaluation combining surrogate metrics, an oracle information-gain measure, and a two-axis \textit{LLM-as-auditor} (\textit{Process Verifiability} and \textit{Evidence Localization}), and show that Chart-RVR produces markedly more verifiable and better-localized rationales than baselines.
\end{itemize}
\section{Related Work}
\noindent \textbf{Chart Reasoning.}
Chart reasoning has been an active area of research recently. Benchmarks for studying chart-related downstream tasks, such as chart-to-table conversion, chart captioning, chart factoid-based question answering, etc. are widely used to evaluate VLMs. Multiple chart-specific VLMs have been proposed, such as Unichart \citep{masry2023unichart}, MatCha \citep{liu2023matcha}, Pix2Struct \citep{lee2023pix2struct}, etc., with considerable success in some of the downstream tasks. However, most of the proposed models struggle when the complexity of the questions increases, which requires relatively deeper reasoning. Some new benchmarks \citep{hegde2025chartqa, ma2025sci} have been proposed to measure both reasoning and accuracy performance in tandem. As a consequence, newer chart reasoning models such as Chartgemma \citep{masry2024chartgemma}, TinyChart \citep{zhang2024tinychart}, ChartAssistant \citep{meng2024chartassistant},
ChartBench \citep{xu2023chartbench}, ChartInsights \citep{wu2024chartinsights}, etc., have been proposed to output rationales with their predictions. Newer approaches utilize contrastive learning \cite{dai2025graph} or use visual tools \cite{fu2025refocus}.

\noindent \textbf{Chain-of-thought in LVLMs.}
Chain-of-thought entails prompting LLMs to think step by step before outputting the final prediction, and provides improvement in both performance and interpretability of LLMs through explicit natural-language reasoning traces \citep{Wei2022CoT}. However, similar observations are not found in LVLMs, where CoT-type prompting significantly degrades performance \citep{zhang2024improve}, especially in smaller models. Several approaches have attempted to improve CoT in LVLMs with further pre-training \citep{xu2024llava}, reinforcement learning \citep{zhang2024improve, xie2024v, liu2025visual}, etc. The degraded CoT performance also hinders the explainability of LVLMs \cite{jiaqi2025think,sinha2026attention}.

% \noindent \textbf{Reinforcement Learning with Verifiable Rewards and GRPO for LVLMs} 
% % waiting for space allowance
% \cite{huang2025vision}

% Sample difficulty \cite{kan2025taco}

% \noindent\textbf{Comparisons to Related Work.}
% Concurrently, BigCharts-R1 introduces a large-scale chart-reasoning dataset constructed via web scraping and trains Qwen-family LVLMs using standard GRPO on this corpus. Our contribution is orthogonal: rather than proposing a new dataset or a Qwen-specific training run, we present a model- and architecture-agnostic fine-tuning recipe that improves chart reasoning via chart-specific verifiable rewards across multiple LVLM families. Moreover, our evaluation targets real-world chart benchmarks (EvoChart, ChartQAPro, ChartBench), whereas BigCharts-R1 primarily reports OOD results on synthetic datasets (e.g., FigureQA, DVQA), where shortcuts and templated visual regularities are common. Our work contributes transferable methodology and real-world generalization through verifiable, task-structured optimization. 

% \newpage
% Our motivation is to provide a strong and generalizable fine-tuning methodology that can utilize well-benchmarked datasets (e.g., ChartQA, ChartFC, PlotQA) for LVLM fine-tuning.

\noindent\textbf{Comparison to Related Work.}
% Two recent lines of chart-reasoning work are most relevant. SFT-based explainable models such as ChartGemma~\citep{masry2025chartgemma} curate gold rationales (e.g., executable-Python traces) and imitate them token-by-token; the rationale itself, however, is never an explicit optimization target, so imitation does not ensure that intermediate values
% are read correctly or that the process is independently checkable.
Concurrently, BigCharts-R1~\cite{masrybigcharts} introduces a
large-scale chart-reasoning corpus constructed via web scraping and trains Qwen-family LVLMs with standard, answer-level GRPO. Our contribution is complementary along two axes. First, rather than a new dataset or a backbone-specific run, Chart-RVR is a model- and architecture-agnostic recipe built entirely from pre-existing datasets, yielding consistent gains across three LVLM families (Qwen2.5VL, Gemma3, InternVL3.5). Second, and more fundamentally, BigCharts-R1 rewards only the final answer, whereas Chart-RVR makes the intermediate process a checkable target, decomposing
the rationale into verifiable Structure and Evidence blocks and a
process-conformity Derivation reward. Finally, our out-of-domain evaluation targets in-the-wild benchmarks (EvoChart, ChartQAPro, ChartBench), whereas BigCharts-R1 reports OOD results primarily on synthetic corpora (e.g., FigureQA, DVQA). Hence, Chart-RVR is a transferable, process-supervised methodology that generalizes to realistic charts, rather than a dataset- or backbone-specific result.
\section{Methodology}
\label{sec:method}

\subsection{Problem Setup and Monitorable Decomposition}
Let $\mathcal{D}=\{(x_i,q_i,y_i^{\ast},a_i^{\ast})\}_{i=1}^N$ be a dataset of chart images $x_i\!\in\!\mathcal{X}$, queries $q_i\!\in\!\mathcal{Q}$, ground-truth answers $y_i^{\ast}\!\in\!\mathcal{Y}$, and reference rationales $a_i^{\ast}\!\in\!\mathcal{A}$.
Mirroring the three verifiable blocks from the Introduction, each rationale factorizes as $a_i^{\ast}=(c_i^{\ast},T_i^{\ast},w_i^{\ast})$: the \textit{Structure} block $c_i^{\ast}\!\in\!\mathcal{C}$ (chart type), the \textit{Evidence} block $T_i^{\ast}\!\in\!\mathcal{T}$ (underlying data table), and the \textit{Derivation} block $w_i^{\ast}\!\in\!\mathcal{W}$ (reasoning trace).
A policy $\pi_\theta$ (an LVLM with parameters $\theta$) produces completions $o_i=(\hat a_i,\hat y_i)\sim\pi_\theta(\cdot\mid x_i,q_i)$ with $\hat a_i=(\hat c_i,\hat T_i,\hat w_i)$, where $\hat a_i$ is the full structured rationale and $\hat y_i$ the final answer.

Chart-RVR elicits responses in a three-block monitorable format, as one tagged completion and assigns rewards to each block, so failures can be localized to Structure, Evidence, and Derivation rather than an unformatted opaque trace. \textbf{Structure} and \textbf{Evidence} are explicit ground-truth targets and are cast as two verifiable surrogate sub-tasks, namely - predicting the chart type $\hat c\in\mathcal{C}$ from a fixed type set (conditioning the model on type-specific cues, e.g.\ bar lengths vs.\ pie sectors), and reconstructing the data table $\hat T=(\hat C,\hat R)$ as JSON with \texttt{`columns'} and \texttt{`rows'} (grounding reasoning in explicit tabular structure), since errors in $\hat T$ corrupt the prerequisites for $\hat y$). 
Finally, the Derivation block has no single ground-truth label and is instead shaped by our Process-Conformity signal (Section~\ref{sec:proc}). All rewards are optimized jointly with GRPO.

\subsection{Group Relative Policy Optimization (GRPO)}
GRPO \citep{shao2024deepseekmath, guo2025deepseek} extends PPO \citep{schulman2017proximal} to group-relative advantage estimation with verifiable rewards. It elicits a set of samples forming a rollout group, computes rewards across them, and updates the policy with the objective discussed below.

\noindent\textbf{Rollout groups.}
For each $(x_i,q_i)$ we sample $G$ rollouts $\{o_j\}_{j=1}^G \sim \pi_{\text{old}}(\cdot \mid x_i,q_i)$ from a frozen behavior policy $\pi_{\text{old}}$, each a completion $o_j=(\hat a_j,\hat y_j)$ parsed into its blocks by the schema tags. Let $\mathrm{tok}(o_j)=(z^{(j)}_{1},\ldots,z^{(j)}_{|o_j|})$ be its tokenization and $z^{(j)}_{<t}=(z^{(j)}_1,\ldots,z^{(j)}_{t-1})$ its prefix.

\noindent\textbf{Objective.}
Within each group, absolute rewards $\{R_j\}_{j=1}^G$ are converted to relative advantages $\hat A_j$:
\begin{equation}
\begin{aligned}
\bar R = \frac{1}{G} \sum_{j=1}^G R_j,
& \qquad
s_R^2 = \frac{1}{G}\sum_{j=1}^G (R_j-\bar R)^2,\\
\hat A_j &= \frac{R_j-\bar R}{\max(1,\, s_R)}.
\end{aligned}
\label{eq:group-adv}
\end{equation}
The policy is updated with a clipped surrogate and a token-averaged KL penalty:
\begin{equation}
\small
\begin{aligned}
\mathcal{J}_{\mathrm{GRPO}}(\theta)
&=
\mathbb{E}_{(x_i,q_i,o_j)\sim\mathcal{D},\,\pi_{\mathrm{old}}}
\Big[
L_{\mathrm{clip}}(\theta)
-\beta D_{\mathrm{KL}}(\pi_\theta \Vert \pi_{\mathrm{ref}})
\Big],
\\
& L_{\mathrm{clip}}(\theta) =
\frac{1}{G}\sum_{j=1}^{G}\frac{1}{|o_j|}
\sum_{t=1}^{|o_j|}
\min\Big\{
\rho_{j,t}(\theta)\hat A_j,
\\
&\hspace{3.7em}
\operatorname{clip}\!\big(
\rho_{j,t}(\theta),1-\epsilon,1+\epsilon
\big)\hat A_j
\Big\}.
\end{aligned}
\label{eq:grpo-objective}
\end{equation}
\vspace{-5pt}
\begin{equation*}
\small
\rho_{j,t}(\theta)
=
\frac{
\pi_{\theta}\!\left(
z^{(j)}_t \mid x_i,q_i,z^{(j)}_{<t}
\right)
}{
\pi_{\mathrm{old}}\!\left(
z^{(j)}_t \mid x_i,q_i,z^{(j)}_{<t}
\right)
}.
\label{eq:grpo-ratio}
\end{equation*}
Here $\epsilon>0$ is clip range, $\pi_{\text{ref}}$ the initial reference policy, and $\beta>0$ weights the token-averaged KL.

\subsection{Schema Rewards: Format, Length, and Accuracy}
Three independent rewards enforce output validity and a formatted answer block. The format reward also enforces the block schema that makes each block individually parseable and thus localizable.

\noindent\textbf{Format reward.}
Regex validation of the two-block schema (\texttt{<think>} then \texttt{<answer>}) with nested \texttt{<type>} (chart type) and \texttt{<table>} (JSON table). We set $R_{\text{fmt}}=1$ if all checks pass, else $0$.

\noindent\textbf{Length (sufficiency) reward.}
Overly short traces underspecify the reasoning while overly long ones overthink and hallucinate \citep{liu2025more}; since our traces are conditioned on the type and table first, we reward tokenized lengths $\ell(o_j)$ within thresholds $0<\eta_1\le\eta_2$:
\[
R_{\text{len}} =
1,~~\text{if}~~ \eta_1 \le \ell(o_j) \le \eta_2, \text{otherwise} ~~0.
\]

\noindent\textbf{Answer accuracy.}
Textual answers are normalized by $\mathrm{norm}$ (lowercase, strip trailing symbols) and matched exactly; numeric answers are matched within a scale-invariant tolerance $\tau$ to absorb imprecise post-computation values:
\begin{equation}
R_{\text{acc}} \;=\;
\begin{cases}
\mathbf{1}\{\mathrm{norm}(\hat y)=\mathrm{norm}(y^\star)\}, & \text{textual},\\[2pt]
\mathbf{1}\left\{\frac{|\hat y - y^\star|}{|y^\star|} \le \tau\right\}, & \text{numeric}.
\end{cases}
\label{eq:acc}
\end{equation}
The combined schema reward is $R_{schema} = R_{fmt}+R_{len}+R_{acc}$.

\subsection{Structure and Evidence Rewards}
\label{sec:surrogate}
The structure and evidence sub-blocks extracted from each response are then individually collected. These two sub-blocks are the verifiable core of monitorability - each is checked against ground truth, enabling a monitor to confirm directly what the model perceived. Their combined reward is $R_{\text{surr}}=R_{\text{type}}+R_{\text{table}}$. 

\noindent\textbf{Structure reward (chart type).}
For ground-truth type $c^{\ast}$ and prediction $\hat c$, an exact match after normalization:
\[
R_{\text{type}}(\hat c,c^{\ast})=\,\mathbf 1\!\bigl[\mathrm{norm}(\hat c)=\mathrm{norm}(c^{\ast})\bigr].
\]

\noindent\textbf{Evidence reward (table reconstruction).}
The model emits a JSON table $\hat{T} =(\hat{C},\hat{R})$ against ground truth $T^{\ast}=({C}^{\ast},{R}^{\ast})$, with $C^{\ast}$ the headers and $R^{\ast}$ the rows, of the form ${\{\text{`columns':\{.. , ..\},`rows':\{[..],[..],..,[..]\} \}}}$.
\begin{equation}
\small
R_{\mathrm{table}}(\hat T,T^{*}) =
\begin{cases}
\begin{aligned}
&
\underbrace{
\frac{1}{|C^{*}|}
\sum_{c\in C^{*}}
\mathbf{1}\{c\in \hat C\}
}_{\text{column header accuracy}}
~~~\text{if valid table} \\
&\quad+
\underbrace{
\frac{1}{|R^{*}|}
\sum_{r\in R^{*}}
\frac{1}{|r|}
\sum_{j}
\mathbf{1}\{r_j=\hat r_j\}
}_{\text{cell accuracy}},
\end{aligned}
\\
0~~\text{otherwise}.
\end{cases}
\label{eq:table-reward}
\end{equation}
Each correct header contributes $1/|{C}^{\ast}|$ and each correct cell $1/(|{R}^{\ast}|\!\times|r|)$. A parseable JSON adds a modest $0.5$ (for reward smoothness); an unparseable one yields $R_{\text{table}}\!=\!0$, encouraging syntactic validity.

\subsection{Derivation Reward: Process Conformity}
\label{sec:proc}
The Structure and Evidence rewards discussed above incentivize accurate perception by the agent which is a prerequisite for accurate reasoning. The Derivation block governs \emph{how} the evidence is utilized to compute the final answer. As Answer-only rewards are sparse and gameable, models can guess or retrofit a fluent rationale \citep{lee2025evaluating} around a correct answer, hence undermining the monitorability of the agent. Similarly, \textit{token-level} imitation (SFT) entrenches dataset-specific styles without ensuring a faithful process. We instead reward \emph{procedural alignment} to a gold reasoning skeleton \emph{in embedding space rather than by surface form}. As chart reasoning is a semi-structured problem, the model may phrase steps in the derivation freely, but must gather the right evidence and apply operations in a consistent order. This gives dense, step-localized credit, discourages decorative CoT. Enforcing an algorithmic skeleton rather than a lexical template improves robustness under domain shift. Conformity \emph{shapes} the derivation to be auditable, while the hard Structure, Evidence, and answer rewards remain the bedrock of correctness in the Derivation process.

Given a text-embedding model $\phi$, we score two sentences $a,b$ by a rescaled cosine similarity $s(a,b)=\tfrac{1}{2}\big(1+\cos(\phi(a),\phi(b))\big)\in[0,1]$, and split the derivation into two stages: gathering the appropriate data, then reasoning over it.

\noindent\textbf{Stage 1: Stepwise grounding ($R_{eg}$).}
We split the derivation into steps and score the first $m$ steps $\hat w_{[:m]}$ \emph{stepwise} against the reference $w^\ast_{[:m]}$, giving localized credit for correctly grounded evidence-gathering.

\noindent\textbf{Stage 2: Reasoning alignment ($R_{rs}$).}
The remaining derivation $\hat w_{[m:]}$ is scored as a whole against $w^\ast_{[m:]}$, preventing drift into degenerate traces.
\begin{equation}
\small
     R_{eg} = \tfrac{1}{m}\sum_i^m s(\hat{w}_{[:m](i)},w^{\ast}_{[:m](i)});~~ R_{rs} = s(\hat w_{[m:]},w^{\ast}_{[m:]}).
\end{equation}
The process-conformity reward is $R_{proc} = R_{eg} + R_{rs}$.

\noindent\textbf{Total reward.}
The final reward is a weighted sum of the Schema rewards ($R_{schema}\in[0,3]$), the Structure-and-Evidence rewards ($R_{surr}$), and the Process-Conformity reward ($R_{proc}\in[0,2]$), with $\lambda_1,\lambda_2>0$:
\begin{equation}
    R = R_{schema} + \lambda_1R_{surr} + \lambda_2R_{proc}.
\end{equation}

\subsection{Why Structured Process States Improve Monitorability}
\label{sec:why-structured}
The three blocks are motivated by an elementary information-theoretic
observation. Let $Q$ be the query, $Y$ the answer, and
$Z^\ast=(C^\ast,T^\ast,D^\ast)$ the structured process state - chart type,
underlying table, and derivation. By the chain rule for conditional mutual
information, conditioning the answer on these blocks is non-increasing in
uncertainty,
\begin{align*}
 & H(Y\mid Q)\;\ge\;H(Y\mid Q,C^\ast)\;\ge\; \\
 & H(Y\mid Q,C^\ast,T^\ast)\;\ge\;H(Y\mid Q,Z^\ast),
\end{align*}

and each step is strict exactly when that block carries an answer-relevant information not already present in the previous ones (e.g.,
$I(Y;T^\ast\mid Q,C^\ast)>0$). In the limit, when the answer is a deterministic function of the query, table, and derivation, $Y=f(Q,T^\ast,D^\ast)$, the residual uncertainty vanishes, $H(Y\mid Q,T^\ast,D^\ast)=0$. This motivates rewarding the blocks jointly rather than optimizing the final answer alone:
each block is incentivized precisely to the extent that it supplies answer-relevant evidence. We emphasize that this characterizes the \emph{oracle} state $Z^\ast$; our method emits estimates $\hat Z=(\hat C,\hat T,\hat D)$, and the empirical sections (Sec.~\ref{sec:monitorability}) measure
how faithfully the emitted state recovers this information and how verifiable and localizable it is in practice.

\begin{figure*}[t]
% \vspace{-20pt}
\centering
\begin{subfigure}{\textwidth}
    \centering
    \includegraphics[width=0.87\linewidth]{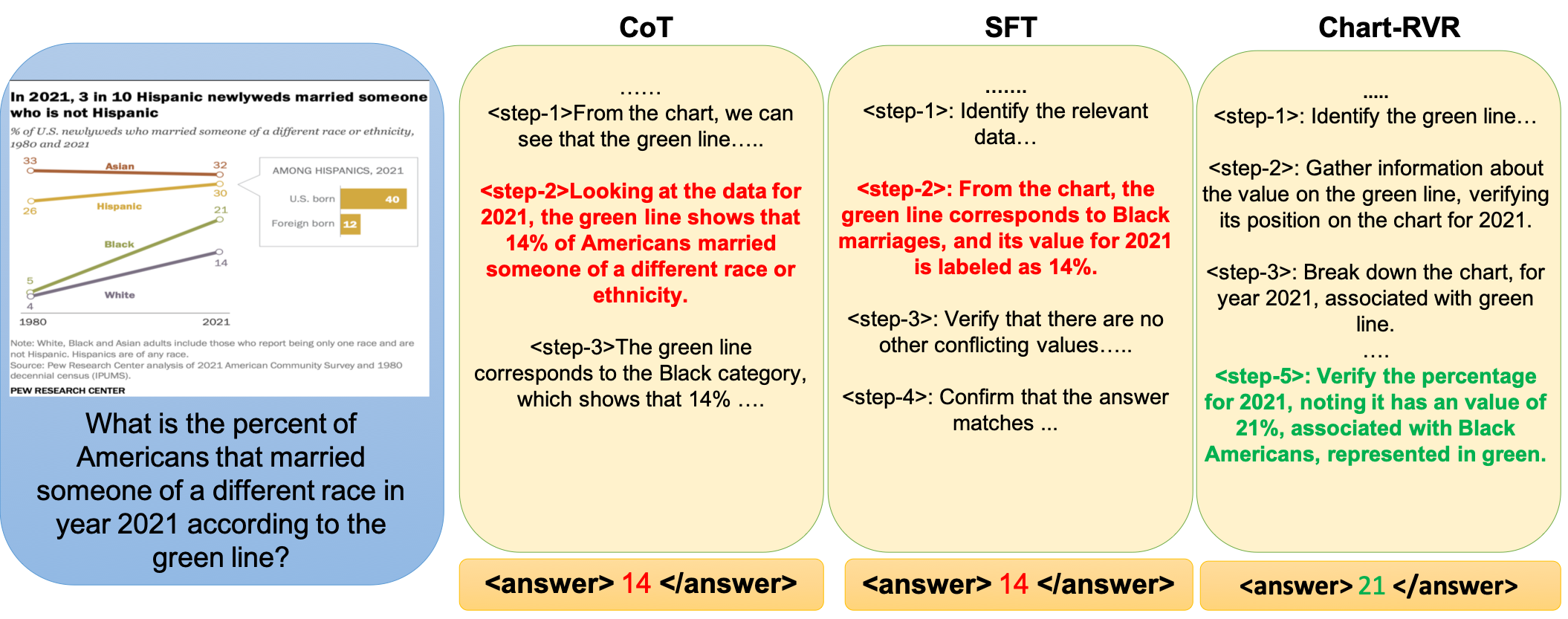}
    \caption{\textbf{Evidence failure: wrong value off the correct series.} All three identify the green (Black-Americans) line, but CoT and SFT extract the wrong value at the evidence-gathering step (\textcolor{red}{red}), reading 14\% instead of 21\%. Notably, SFT's later ``verification'' steps are decorative-they affirm the fabricated value rather than re-reading the chart, illustrating how a fluent, self-confirming rationale conceals its own error. Chart-RVR re-grounds in the chart, recovers the correct value (\textcolor{green}{green}), and confines the decision to a single checkable step.}
    \label{fig:comparision:a}
\end{subfigure}
\vspace{-5pt}
\begin{subfigure}{\textwidth}
    \centering
    \includegraphics[width=0.76\linewidth]{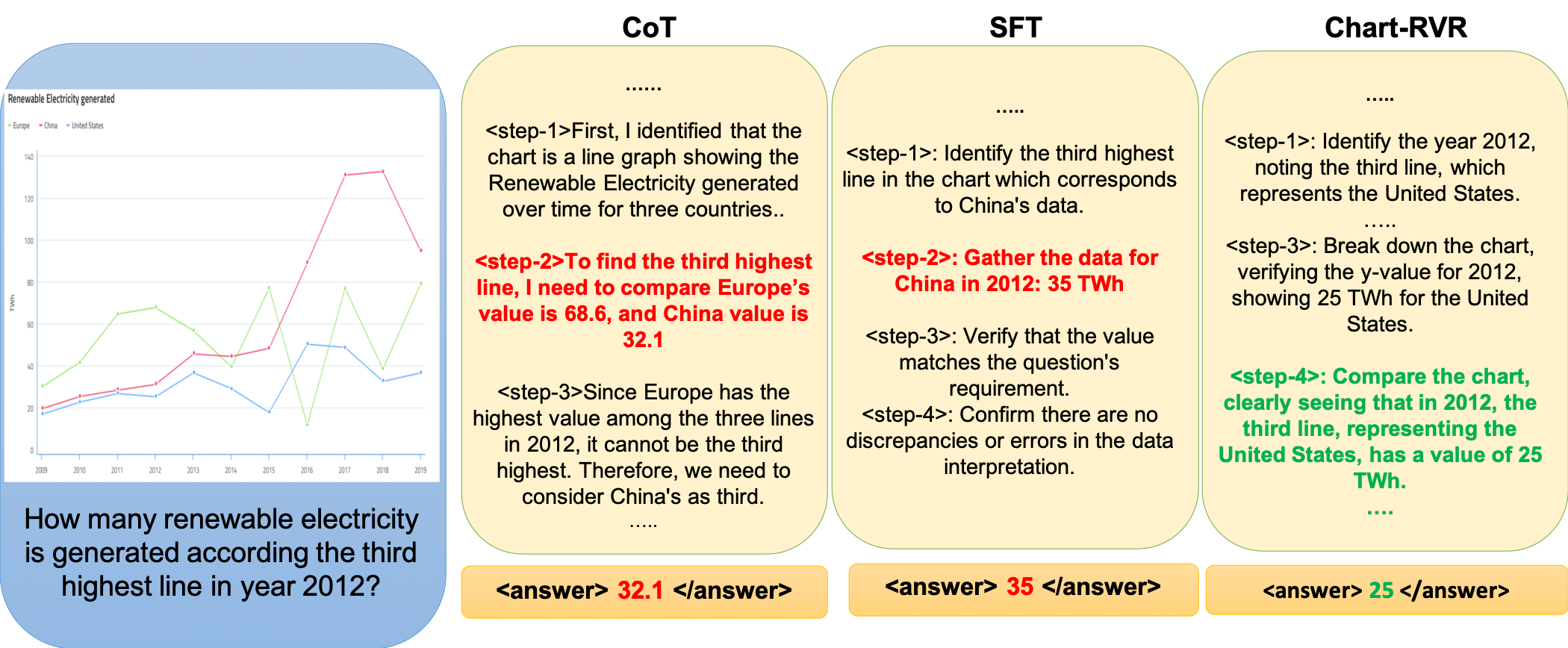}
    \caption{\textbf{Evidence failure: wrong series identified.} A distinct failure mode-selecting which line is third-highest in 2012. CoT picks Europe and SFT picks China (\textcolor{red}{red}), then each confidently computes a value from the wrong series; Chart-RVR correctly localizes the third-highest line to the United States and reads 25\,TWh (\textcolor{green}{green}). Because the error is confined to the series-identification step, an auditor can pinpoint and repair it without re-deriving the trace.}
    \label{fig:comparision:b}
\end{subfigure}
% \vspace{-15pt}
\caption{\textbf{Localizable failures on EvoChart (OOD).} Verbatim rationale excerpts (elisions marked ``\ldots'') for three conditions: the base model with CoT prompting, the SFT model, and Chart-RVR. Erroneous steps are in \textcolor{red}{red}, correctly grounded steps in \textcolor{green}{green}. In both cases the baselines fail at the \textit{Evidence} block-value extraction in (a), series identification in (b)-and the failure pins to a single, identifiable step rather than diffusing across a fluent trace. Crucially, step structure alone is insufficient: the SFT traces are equally step-structured yet still fail, and their verification steps merely rubber-stamp the hallucinated value; Chart-RVR's Structure and Evidence rewards instead force the trace to re-ground in the chart. See Appendix for additional examples.}
\label{fig:comparision}
% \vspace{-15pt}
\end{figure*}

\section{Experiments and Results}
\subsection{Dataset and Model Settings}
\label{sec:dset}
\noindent\textbf{Datasets.} We build our CoT training data from three datasets: ChartQA \citep{masry2022chartqa} (factual and reasoning questions), PlotQA \citep{Methani_2020_WACV} (synthetic factoid questions), and ChartFC \citep{akhtar2023reading} (binary reasoning questions); their test splits serve as our \textbf{in-domain} benchmarks. For \textbf{out-of-domain} evaluation we use EvoChart \citep{huang2025evochart} (irregular in-the-wild charts), ChartQAPro \citep{masry2025chartqapro} (harder charts and questions), ChartBench \citep{xu2023chartbench} (complex reasoning), and the reasoning-only CharXiv \citep{wang2024charxiv}.

\noindent\textbf{Models and baselines.} All main experiments use Qwen2.5VL-3B-Instruct \citep{qwen2.5-VL}; to show generalizability, we also train Gemma3-3B-it \citep{team2025gemma} and InternVL3.5-4B \citep{wang2025internvl3}. We compare against ChartGemma \citep{masry2024chartgemma}, a PaliGemma-based model emitting executable-Python rationales and the strongest explainable baseline at 3-4B scale. We further compare against BigCharts-R1 \citep{masrybigcharts}; as neither its data nor models are public, we reimplement their reward recipe faithfully and train it on the \emph{same backbone and data} as ours (the \texttt{Q2.5VL-SFT+GRPO} row), isolating the comparison to the reward design.

\noindent\textbf{Evaluation.} We report Relaxed Accuracy \citep{masry2022chartqa}, counting a numeric prediction correct within a $5\%$ tolerance, 
\[
\texttt{RelAcc}(\hat{y},y^{\ast}) = \mathbf{1}\frac{|y^{\ast}-\hat{y}|}{|y^{\ast}|}[ \le 0.05]~~\text{for}~y^{\ast}\neq 0
\]
matching the tolerance $\tau$ of our accuracy reward (Equation~\ref{eq:acc}); non-numeric answers use exact match. For ChartFC, since some semantically correct ``True/False'' replies are mis-scored under Yes/No matching, we append ``Answer Yes/No'' to the prompt.

\noindent\textbf{Hyperparameters.} The process-conformity step count is $m=3$ (the average trace length), and the length-reward band is $[\eta_1,\eta_2]=[150,400]$ tokens. Remaining settings and reward curves are in the Appendix.

\subsection{Structured Prompt Evaluation}  
First, we evaluated how chain-of-thought prompting affects off-the-shelf LVLMs in Table~\ref{tab:cot-vs-structure-cot}. Although CoT prompting shows gains in LLMs, we observed \citep{liu2025visual,xu2024llava} an opposite trend in LVLMs, where standard CoT prompting degrades performance by a large margin compared to direct prompting. Next, we examined how structured prompting (i.e., instructing the model to emit chart type, table, and the reasoning process along with the answer) using the prompt structure shown in Figure~\ref{fig:grpo-prompt} improves the results over standard CoT prompting. As can be seen, our structured prompting approach improved performance, but is still significantly less than direct prompting. 

\begin{table}[h]
% \vspace{-40pt}
\centering
\caption{Comparison of Direct, CoT, and Structured Prompt on off-the-shelf Qwen2.5VL-3B-Instruct. Our prompt template is detailed in Figure~\ref{fig:grpo-prompt}.}
% \vspace{-10pt}
\resizebox{0.9\linewidth}{!}{\begin{tabular}{lccc}
\toprule
\textbf{Dataset} & \textbf{Direct} & \textbf{CoT} & \textbf{Structured} \\
\midrule
\bf ChartQA & 82.0 & 41.8 & \bf 73.12  \\
\bf PlotQA & 80.5 & 31.82 & \bf 52.72  \\
\bf ChartFC & 74.4 & 48.02 & \bf 69.20 \\
\midrule
\bf EvoChart & 48.72 & 18.72 & \bf 29.60 \\
\bf ChartQAPro & 25.7 & 12.01 & \bf 15.80 \\
\bf ChartBench & 66.04 & 29.4 & \bf 51.16 \\
\bottomrule
\end{tabular}}
% \vspace{-10pt}
\label{tab:cot-vs-structure-cot}
\end{table} 

\begin{table*}[th]
    \centering 
    \caption{Main benchmark results across 6 diverse chart datasets. The `Exp?' column signifies if the approach is explainable (i.e., outputs CoT rationales or equivalent, like Python programs). The performance improvement is more pronounced on out-of-domain (OOD) datasets (last 3 columns).}
  \setlength{\tabcolsep}{4pt} 
  \resizebox{0.95\textwidth}{!}{
  \begin{tabular}{@{}l*{8}{c}@{}}
    \toprule
    \textbf{Approach} & \textbf{Exp?} & \textbf{ChartQA} & \textbf{PlotQA} & \textbf{ChartFC} &
    \textbf{EvoChart} & \textbf{ChartQAPro} & \textbf{ChartBench} \\
    \midrule
    \multicolumn{8}{@{}l@{}}{\textbf{Direct Prompting}} \\
    \midrule
    Q2.5VL-Ins & \xmark & 82.0 & 80.5 & 74.4 & 48.72 & 25.7 &  66.04  \\
    \midrule
    \multicolumn{8}{@{}l@{}}{\textbf{Explainable Models with Rationales}} \\
    \midrule
    Q2.5VL-Ins (CoT) & \redcheck & 73.12 & 52.72  & 69.20 &  29.6 & 15.80 &  51.16  \\
    ChartGemma \cite{masry2024chartgemma} & \redcheck & 76.44 & 33.28  & 70.33 & 36.96 & 10.93 &  40.56   \\
    \midrule
    \multicolumn{8}{@{}l@{}}{\textbf{Fine-tuned Models with Rationales}} \\
    \midrule
    Q2.5VL-SFT & \redcheck & 83.08 & 74.18 & 77.30  &  46.08 & 23.56  & 64.64  \\
    Q2.5VL-SFT + GRPO \cite{masrybigcharts} & \redcheck &  84.18 & 77.51 & 77.45 & 45.14  & 24.92  & 66.32 \\
    Q2.5VL-Ins (A+F+L) & \redcheck & 76.72 & 56.22 & 58.58 & 38.88 & 17.55 & 48.1  \\
    Q2.5VL-Ins (A+F+L+Tasks) & \redcheck & 81.8 & 76.24 & 63.85 & 51.68 & 27.66  & 65.28  \\
    \textbf{Chart-RVR-3B} (Ours) & \redcheck & 84.56 & \textbf{78.68} & 77.62 & 53.36 & 28.38  & 68.32  \\
    \midrule
    \multicolumn{8}{@{}l@{}}{\textbf{Curated Data Fine-tuned Models with Rationales}} \\
    \midrule
    Q2.5VL-SFT-Hard & \redcheck & 84.28 & 75.54 & 77.90  &  49.36 &  23.20 &  65.12  \\
    \textbf{Chart-RVR-3B-Hard} (Ours) & \redcheck & \textbf{85.76} & 77.9 & \textbf{80.07} & \textbf{54.24} & \textbf{28.64} &  \textbf{69.46} \\
    \bottomrule
  \end{tabular}}
  \label{tab:chart_rvr_large}
\end{table*}

\subsection{Benchmark Performance}
Table~\ref{tab:chart_rvr_large} reports accuracy on six benchmarks. Chart-RVR outperforms all explainable baselines - including ChartGemma and the BigCharts-R1 reward recipe and uniquely among explainable methods, \emph{exceeds non-explainable direct prompting} (85.76 vs.\ 82.0 on ChartQA), whereas CoT and ChartGemma fall well below it. Gains are observed in in-domain 1--3\% and OOD datasets EvoChart \textbf{+7.28\%}, ChartQAPro \textbf{+4.82\%}, ChartBench \textbf{+3.68\%}. The Hard variant adds a further 1-2\% gains, implying the effectiveness of data curation. (Note: PlotQA is not reasoning-heavy). On the significantly hard OOD reasoning-only dataset - CharXiv (Table~\ref{tab:charxiv-reasoning}), Chart-RVR-Hard improves \textbf{+4.7\%} over SFT (Note the accuracy is adjudged by LLM judge as defined in CharXiv protocol).

\noindent\textbf{Reward ablation.} The rows in Table~\ref{tab:chart_rvr_large} isolate each reward group. Standard GRPO (A+F+L) underperforms SFT - \textbf{naive GRPO is ineffective}; the surrogate-task rewards (A+F+L+Tasks) recover and surpass it OOD; and the process-conformity reward delivers the final, consistent gains across all benchmarks, confirming its effectiveness over generic GRPO.

\begin{table}[h]
\centering
\caption{Performance on the CharXiv reasoning subset.}
% \vspace{-10pt}
\resizebox{0.8\linewidth}{!}{
\begin{tabular}{lc}
\toprule
Method    & CharXiv (reasoning) \\
\midrule
CoT       & 26.2 \\
Q2.5VL-SFT-Hard       & 29.6 \\
Chart-RVR-3B-Hard & \textbf{34.3} \\
\bottomrule
\end{tabular}}
\label{tab:charxiv-reasoning}
% \vspace{-15pt}
\end{table}

\subsection{Robustness across Architectures and Seeds}
To confirm the gains are not artifacts of a single split or backbone, across backbones (Table~\ref{tab:ablations-training-settings}), Chart-RVR yields consistent OOD improvements on both Gemma3 and InternVL3.5, where naive GRPO+Tasks is inconsistent, trailing only on InternVL ChartQAPro (at par), evidencing architecture-agnostic generalization. Next, we (i) retrain on three random 6k subsets of ChartQA/PlotQA/ChartFC. Across seeds, Chart-RVR beats SFT by \textbf{1--2\%} ID and \textbf{4--6\%} OOD with low variance (Table~\ref{tab:sft-vs-rvr}) (Appendix), matching the single-split results and indicating robust convergence.

\begin{table}[h]
\centering
\setlength{\tabcolsep}{2pt}
\caption{\textbf{Ablation across baselines on OOD benchmarks.} Subcolumns are Gemma-3 and InternVL-3.5 (IVL). Chart-RVR outperforms across datasets.}
\resizebox{1\linewidth}{!}{%
\begin{tabular}{l*{6}{c}}
\toprule
\multicolumn{1}{c}{} &
\multicolumn{2}{c}{\ds{EvoChart}} &
\multicolumn{2}{c}{\ds{ChartQAPro}} &
\multicolumn{2}{c}{\ds{ChartBench}} \\
\cmidrule(lr){2-3} \cmidrule(lr){4-5} \cmidrule(lr){6-7} 
\textbf{Setting} &
\textbf{Gemma} & \textbf{IVL} &
\textbf{Gemma} & \textbf{IVL} &
\textbf{Gemma} & \textbf{IVL} \\
\midrule 
CoT & 21.04 & 45.36 & 24.82 & 15.91 & 42.28 & 54.42\\ 
SFT  &  21.92 & 44.08  & 23.56  & \best{29.34}  & 54.14  & 63.16 \\
GRPO (A+F+L+Tasks) & 42.26 & 50.04 & 29.51 & 20.43 & 53.18 & 64.62 \\
\textbf{Chart-RVR (Ours)} &  \best{42.48} & \best{50.24} &  \best{32.59} & \best{29.34} &  \best{58.18} & \best{64.78} \\
\bottomrule
\end{tabular}%
}
\label{tab:ablations-training-settings}
% \vspace{-10pt}
\end{table}

\subsection{Monitorability}
\label{sec:monitorability}
Beyond accuracy, we assess whether rationales are \emph{monitorable} via a triangulated protocol covering the three properties: \emph{grounded} metrics computed against ground-truth/oracle targets (no LLM judgment) for \textbf{(1) verifiability} and \textbf{(2) localizability}; an \emph{LLM-as-auditor}, validated against those grounded signals, for \textbf{(3) auditability}; and an oracle information-gain measure of rationale \textbf{(4) usefulness}.

\noindent\textbf{(1) Verifiability (grounded).}
Structure accuracy $\mathrm{Acc}_c=\mathbf 1[\mathrm{norm}(\hat c)=\mathrm{norm}(c^{\ast})]$ and table fidelity (normalized edit distance $\mathrm{ED}$, Eq.~\ref{eq:editdist}) measure whether the Structure and Evidence blocks check out against ground truth (Table~\ref{tab:surrogate}).
\begin{equation}
\small
    E_{T}(\hat T,T^{\ast}) = \tfrac{1}{|{C}^{\ast}|} \!\sum_{c \in C^*}\! \mathbf{1}[c \notin \hat{C}] + \tfrac{1}{|{R}^{\ast}|r} \!\sum_{r \in R^*}\sum_{j}\! \mathbf{1}[r_j \neq \hat{r}_j].
\label{eq:editdist}
\end{equation}
As shown in Table~\ref{tab:surrogate}, SFT and Chart-RVR both lift chart-type accuracy (the base model is already decent), while Chart-RVR improves table reconstruction on EvoChart (OOD) by $0.06$ over SFT - explicitly grounding reasoning in verifiable data.

\begin{table}[h]
\centering
\caption{Verifiability and HR metrics: Chart Type Accuracy (Acc, $\uparrow$) and table reconstruction (Tab, edit-distance $\downarrow$). Last column represents Hallucination Rate (HR, $\downarrow$). Dataset order follows Table~\ref{tab:chart_rvr_large}.}
\resizebox{\linewidth}{!}{
\begin{tabular}{l|ccc|ccc|ccc}
\toprule
\textbf{Dset} & \multicolumn{3}{c|}{\textbf{CoT}} & \multicolumn{3}{c|}{\textbf{SFT}} & \multicolumn{3}{c}{\textbf{Chart-RVR}} \\
\midrule
 & \bf Acc  & \bf Tab  & \bf HR  & \bf Acc  & \bf Tab  & \bf HR  & \bf Acc & \bf Tab  & \bf HR   \\
\midrule
CQA   & 0.87 & 0.46 & 0.15 & 0.94 & \bf 0.38 & 0.11  & \bf 0.95 & 0.49 & \bf 0.10 \\
PQA  & 0.70 & 0.65 & 0.12 & \bf 0.78 & 1.13 & 0.08 & 0.77 & \bf 1.03 & \bf 0.08  \\
CFC  & 1.00 & 0.65 & 0.18 & 1.00 & 0.20 & 0.14 & \bf 1.00  & \bf 0.20 & \bf 0.13 \\
\midrule
EC  & 0.74 &  0.99 & 0.31 & 0.81 & 1.28 & 0.26 & \bf 0.84 & \bf 1.22 & \bf 0.18 \\
CPro & 0.69 & 0.72 & 0.46 & 0.70 & 1.05 & 0.41 & \bf 0.72 & \bf 1.03 & \bf 0.34 \\
\bottomrule
\end{tabular}}
\label{tab:surrogate}
\vspace{-10pt}
\end{table}

\noindent\textbf{(2) Localizability.}
To compute the fidelity of each evidence gathering step, we reuse the tolerant matching of numerical values in the first $m$ steps i.e. the set $\mathcal{V}(\hat w)$ of numeric quantities cited in the derivation (extracted by a deterministic parser), a value $v$ is \emph{supported} when some cell $u$ of the reference table $T^{\ast}$ lies within relative tolerance $\tau$,
\[
\mathrm{match}(v,T^{\ast})=\mathbf 1\!\Big[\exists\,u\in T^{\ast}:\ \tfrac{|v-u|}{\max(|u|,\varepsilon)}\le\tau\Big],
\]
where $\varepsilon$ is a small constant guarding against division by zero, grounding precision is calculated as $\mathrm{GP}(\hat w)=|\mathcal{V}|^{-1}\sum_{v\in\mathcal{V}}\mathrm{match}(v,T^{\ast})$ and hallucination rate $\mathrm{HR}=1-\mathrm{GP}$ then measure how many cited values are genuinely traceable to the chart - under the same $5\%$ band used for answer scoring, so a step counts as grounded iff its number is as close to a real data point as a correct answer must be. Chart-RVR attains the highest $\mathrm{GP}$ and lowest $\mathrm{HR}$, with the largest margins on the OOD benchmarks (Table~\ref{tab:surrogate}).

\noindent\textbf{Auditability (LLM-as-auditor).}
An oracle $\mathcal{O}\in\{\text{GPT-4o},\text{LLaVA-Next-72B}\}$, given $(x,q,\hat a)$, scores \emph{Process Verifiability} (PV: is the trace decomposed into checkable steps stating intermediate quantities?) and \emph{Evidence Localization} (EL: is each step tied to a specific chart/table element?) on $1$--$5$ (rubric in Appendix). Chart-RVR scores highest on both axes under both oracles (Table~\ref{tab:auditor}), with the $\text{+Tasks}\!\rightarrow\!\text{Ours}$ jump isolating process-conformity as the driver.
\begin{table}[h]
\centering
\small
\caption{\textbf{LLM-as-auditor scores} (1--5 $\uparrow$), averaged over six benchmarks under two oracles. \textbf{PV}: Process Verifiability; \textbf{EL}: Evidence Localization. The ``Surrogate (no PC)'' row is the \texttt{A+F+L+Tasks} ablation; the gain to ours isolates the process-conformity (PC) reward.}
\setlength{\tabcolsep}{5pt}
\renewcommand{\arraystretch}{1.05}
\resizebox{\linewidth}{!}{
\begin{tabular}{lcccc}
\toprule
& \multicolumn{2}{c}{\textbf{GPT-4o}} & \multicolumn{2}{c}{\textbf{LLaVA-Next-72B}} \\
\cmidrule(lr){2-3}\cmidrule(lr){4-5}
\textbf{Method} & \textbf{PV}\,$\uparrow$ & \textbf{EL}\,$\uparrow$ & \textbf{PV}\,$\uparrow$ & \textbf{EL}\,$\uparrow$ \\
\midrule
CoT prompting              & 2.41 & 2.18 & 2.55 & 2.30 \\
SFT                        & 3.12 & 2.96 & 3.20 & 3.05 \\
Surrogate (no PC)   & 3.58 & 3.41 & 3.63 & 3.49 \\
\textbf{Chart-RVR (Ours)}  & \textbf{4.36} & \textbf{4.28} & \textbf{4.41} & \textbf{4.33} \\
\bottomrule
\end{tabular}}

\label{tab:auditor}
\vspace{-10pt}
\end{table}

\noindent\textbf{Usefulness ($\Delta\log P$).}
Finally, the Explainable Information Gain $\Delta\log P = \log P_{\mathcal{O}}(y^{\ast}\!\mid\! x,a)-\log P_{\mathcal{O}}(y^{\ast}\!\mid\! x)$ measures how much a rationale raises an oracle's certainty in the correct answer \citep{deyoung2020eraser,tutek2025measuring,paul2024making} (Qwen2.5VL-72B oracle). Table~\ref{tab:explanation-metric} shows CoT rationales \emph{reduce} certainty (negative), while Chart-RVR \emph{adds} information, especially OOD (EvoChart $+0.02$ vs.\ CoT $-9.13$; PlotQA $+3.66$; ChartQAPro $+2.41$). All methods dip on ChartQA (base-model memorization) but improve on the harder ChartQA-human split. A second oracle (LLaVA-Next-72B, App.~Table~\ref{tab:explanation-metric-appendix}) and a human study (App.~Fig.~\ref{fig:human-study}) corroborate, and Figure~\ref{fig:comparision} gives a qualitative example.

\begin{table}[h]
\centering
\caption{Explainable Information Gain ($\Delta\log P$) on `Hard' variants; positive means the rationale raises certainty in the correct answer. We observe identical result trends on LLaVA-72B in the Appendix.}
% \vspace{-5pt}
\resizebox{0.85\linewidth}{!}{
\begin{tabular}{c|c|c|c}
\toprule
\textbf{Dataset} & \textbf{$\Delta$ CoT} & \textbf{$\Delta$ SFT} & \textbf{$\Delta$ Chart-RVR} \\
\midrule
 ChartQA & -5.04 & \bf -2.22 & -4.3 \\
 ChartQA (human) & -3.46 & +0.09 & \bf +0.3 \\
 PlotQA & -2.25 & +1.13 & \bf +3.66 \\
 \midrule
 EvoChart & -9.13 & -6.82 & \bf +0.02  \\
 ChartQAPro & -0.04 & +1.75  & \bf +2.41 \\
 \bottomrule
\end{tabular}}
\label{tab:explanation-metric}
% \vspace{-10pt}
\end{table}

\section{Conclusion}
We introduced Chart-RVR, a reinforcement learning framework that turns chart reasoning into a monitorable process by augmenting GRPO with verifiable rewards over chart-type prediction and table reconstruction, together with a process-conformity objective that shapes faithful, step-by-step derivations. Across six benchmarks and three LVLMs, Chart-RVR improves accuracy while producing markedly more verifiable and better-localized rationales, with the largest gains under distribution shift - evidence of stronger OOD generalization than SFT and vanilla GRPO. 
% \newpage
\section*{Acknowledgments}
We are thankful to CoreWeave, whose purpose-built AI cloud platform powered our experiments. We are thankful to Morgan Stanley for providing research mentorship. A part of this work is also supported by the US National Science Foundation (NSF) under grants IIS-2106913, IIS-2538206, IIS-2529378, CCF-2217071, and CNS-2213700. Any recommendations expressed in this material are those of the authors and do not necessarily reflect the views of NSF. OpenAI's ChatGPT and Anthropic's Claude were used to assist with language editing of this manuscript. All generated content was reviewed and verified by the authors, who assume full responsibility for the final manuscript.

\section*{Limitations}
Our study targets small (3-4B) LVLMs across three backbones; gains at larger closed-source model scales are untested. Our LLM-as-auditor axes inherit judge subjectivity, mitigated by two oracles, grounded-signal corroboration, and a (small-scale) human study. We evaluate only English ChartQA and leave a rigorous \textit{Intervention Utility} protocol, repairing localized failures from the exposed process state to future work.

\bibliography{custom}

@article{xu2024llava,
  title={Llava-o1: Let vision language models reason step-by-step},
  author={Xu, Guowei and Jin, Peng and Hao, Li and Song, Yibing and Sun, Lichao and Yuan, Li},
  journal={arXiv preprint arXiv:2411.10440},
  year={2024}
}

@inproceedings{islam2024large,
    title = "Are Large Vision Language Models up to the Challenge of Chart Comprehension and Reasoning",
    author = "Islam, Mohammed Saidul  and
      Rahman, Raian  and
      Masry, Ahmed  and
      Laskar, Md Tahmid Rahman  and
      Nayeem, Mir Tafseer  and
      Hoque, Enamul",
    editor = "Al-Onaizan, Yaser  and
      Bansal, Mohit  and
      Chen, Yun-Nung",
    booktitle = "Findings of the Association for Computational Linguistics: EMNLP 2024",
    month = nov,
    year = "2024",
    address = "Miami, Florida, USA",
    publisher = "Association for Computational Linguistics",
    url = "https://aclanthology.org/2024.findings-emnlp.191/",
    doi = "10.18653/v1/2024.findings-emnlp.191",
    pages = "3334--3368",
}

@inproceedings{masry2024chartgemma,
    title = "{C}hart{G}emma: Visual Instruction-tuning for Chart Reasoning in the Wild",
    author = "Masry, Ahmed  and
      Thakkar, Megh  and
      Bajaj, Aayush  and
      Kartha, Aaryaman  and
      Hoque, Enamul  and
      Joty, Shafiq",
    editor = "Rambow, Owen  and
      Wanner, Leo  and
      Apidianaki, Marianna  and
      Al-Khalifa, Hend  and
      Eugenio, Barbara Di  and
      Schockaert, Steven  and
      Darwish, Kareem  and
      Agarwal, Apoorv",
    booktitle = "Proceedings of the 31st International Conference on Computational Linguistics: Industry Track",
    month = jan,
    year = "2025",
    address = "Abu Dhabi, UAE",
    publisher = "Association for Computational Linguistics",
    url = "https://aclanthology.org/2025.coling-industry.54/",
    pages = "625--643",
}

@inproceedings{
fu2025refocus,
title={ReFocus: Visual Editing as a Chain of Thought for Structured Image Understanding},
author={Xingyu Fu and Minqian Liu and Zhengyuan Yang and John Richard Corring and Yijuan Lu and Jianwei Yang and Dan Roth and Dinei Florencio and Cha Zhang},
booktitle={Forty-second International Conference on Machine Learning},
year={2025},
url={https://openreview.net/forum?id=a7qFlPOTix}
}

@inproceedings{carbune2024chart,
    title = "Chart-based Reasoning: Transferring Capabilities from {LLM}s to {VLM}s",
    author = "Carbune, Victor  and
      Mansoor, Hassan  and
      Liu, Fangyu  and
      Aralikatte, Rahul  and
      Baechler, Gilles  and
      Chen, Jindong  and
      Sharma, Abhanshu",
    editor = "Duh, Kevin  and
      Gomez, Helena  and
      Bethard, Steven",
    booktitle = "Findings of the Association for Computational Linguistics: NAACL 2024",
    month = jun,
    year = "2024",
    address = "Mexico City, Mexico",
    publisher = "Association for Computational Linguistics",
    url = "https://aclanthology.org/2024.findings-naacl.62/",
    doi = "10.18653/v1/2024.findings-naacl.62",
    pages = "989--1004",
}

@inproceedings{xie2024v,
    title = "{V}-{DPO}: Mitigating Hallucination in Large Vision Language Models via Vision-Guided Direct Preference Optimization",
    author = "Xie, Yuxi  and
      Li, Guanzhen  and
      Xu, Xiao  and
      Kan, Min-Yen",
    editor = "Al-Onaizan, Yaser  and
      Bansal, Mohit  and
      Chen, Yun-Nung",
    booktitle = "Findings of the Association for Computational Linguistics: EMNLP 2024",
    month = nov,
    year = "2024",
    address = "Miami, Florida, USA",
    publisher = "Association for Computational Linguistics",
    url = "https://aclanthology.org/2024.findings-emnlp.775/",
    doi = "10.18653/v1/2024.findings-emnlp.775",
    pages = "13258--13273",
}

@article{liu2025visual,
  title={Visual-rft: Visual reinforcement fine-tuning},
  author={Liu, Ziyu and Sun, Zeyi and Zang, Yuhang and Dong, Xiaoyi and Cao, Yuhang and Duan, Haodong and Lin, Dahua and Wang, Jiaqi},
  journal={arXiv preprint arXiv:2503.01785},
  year={2025}
}

@inproceedings{huang2025evochart,
  title={Evochart: A benchmark and a self-training approach towards real-world chart understanding},
  author={Huang, Muye and Lai, Han and Zhang, Xinyu and Wu, Wenjun and Ma, Jie and Zhang, Lingling and Liu, Jun},
  booktitle={Proceedings of the AAAI Conference on Artificial Intelligence},
  volume={39},
  number={4},
  pages={3680--3688},
  year={2025}
}

@article{guo2025deepseek,
  title={Deepseek-r1: Incentivizing reasoning capability in llms via reinforcement learning},
  author={Guo, Daya and Yang, Dejian and Zhang, Haowei and Song, Junxiao and Zhang, Ruoyu and Xu, Runxin and Zhu, Qihao and Ma, Shirong and Wang, Peiyi and Bi, Xiao and others},
  journal={arXiv preprint arXiv:2501.12948},
  year={2025}
}

@inproceedings{masry2025chartqapro,
    title = "{C}hart{QAP}ro: A More Diverse and Challenging Benchmark for Chart Question Answering",
    author = "Masry, Ahmed  and
      Islam, Mohammed Saidul  and
      Ahmed, Mahir  and
      Bajaj, Aayush  and
      Kabir, Firoz  and
      Kartha, Aaryaman  and
      Laskar, Md Tahmid Rahman  and
      Rahman, Mizanur  and
      Rahman, Shadikur  and
      Shahmohammadi, Mehrad  and
      Thakkar, Megh  and
      Parvez, Md Rizwan  and
      Hoque, Enamul  and
      Joty, Shafiq",
    editor = "Che, Wanxiang  and
      Nabende, Joyce  and
      Shutova, Ekaterina  and
      Pilehvar, Mohammad Taher",
    booktitle = "Findings of the Association for Computational Linguistics: ACL 2025",
    month = jul,
    year = "2025",
    address = "Vienna, Austria",
    publisher = "Association for Computational Linguistics",
    url = "https://aclanthology.org/2025.findings-acl.978/",
    doi = "10.18653/v1/2025.findings-acl.978",
    pages = "19123--19151",
    ISBN = "979-8-89176-256-5",
}

@article{xu2023chartbench,
  title={Chartbench: A benchmark for complex visual reasoning in charts},
  author={Xu, Zhengzhuo and Du, Sinan and Qi, Yiyan and Xu, Chengjin and Yuan, Chun and Guo, Jian},
  journal={arXiv preprint arXiv:2312.15915},
  year={2023}
}

@article{qwen2.5-VL,
  title={Qwen2.5-VL Technical Report},
  author={Bai, Shuai and Chen, Keqin and Liu, Xuejing and Wang, Jialin and Ge, Wenbin and Song, Sibo and Dang, Kai and Wang, Peng and Wang, Shijie and Tang, Jun and Zhong, Humen and Zhu, Yuanzhi and Yang, Mingkun and Li, Zhaohai and Wan, Jianqiang and Wang, Pengfei and Ding, Wei and Fu, Zheren and Xu, Yiheng and Ye, Jiabo and Zhang, Xi and Xie, Tianbao and Cheng, Zesen and Zhang, Hang and Yang, Zhibo and Xu, Haiyang and Lin, Junyang},
  journal={arXiv preprint arXiv:2502.13923},
  year={2025}
}

@inproceedings{masry2022chartqa,
  title={ChartQA: A Benchmark for Question Answering about Charts with Visual and Logical Reasoning},
  author={Masry, Ahmed and Do, Xuan Long and Tan, Jia Qing and Joty, Shafiq and Hoque, Enamul},
  booktitle={Findings of the Association for Computational Linguistics: ACL 2022},
  pages={2263--2279},
  year={2022}
}

@misc{vonwerra2022trl,
    title        = {{TRL: Transformer Reinforcement Learning}},
    author       = {Leandro von Werra and Younes Belkada and Lewis Tunstall and Edward Beeching and Tristan Thrush and Nathan Lambert and Shengyi Huang and Kashif Rasul and Quentin Gallou{\'e}dec},
    year         = 2020,
    journal      = {GitHub repository},
    publisher    = {GitHub},
    howpublished = {\url{https://github.com/huggingface/trl}}
}

@InProceedings{Methani_2020_WACV,
author = {Methani, Nitesh and Ganguly, Pritha and Khapra, Mitesh M. and Kumar, Pratyush},
title = {PlotQA: Reasoning over Scientific Plots},
booktitle = {The IEEE Winter Conference on Applications of Computer Vision (WACV)},
month = {March},
year = {2020}
}

@inproceedings{akhtar2023reading,
  title={Reading and Reasoning over Chart Images for Evidence-based Automated Fact-Checking},
  author={Akhtar, Mubashara and Cocarascu, Oana and Simperl, Elena},
  booktitle={Findings of the Association for Computational Linguistics: EACL 2023},
  pages={399--414},
  year={2023}
}

@inproceedings{masry2023unichart,
  title={UniChart: A Universal Vision-language Pretrained Model for Chart Comprehension and Reasoning},
  author={Masry, Ahmed and Kavehzadeh, Parsa and Do, Xuan Long and Hoque, Enamul and Joty, Shafiq},
  booktitle={Proceedings of the 2023 Conference on Empirical Methods in Natural Language Processing},
  pages={14662--14684},
  year={2023}
}

@inproceedings{liu2023matcha,
  title={MatCha: Enhancing Visual Language Pretraining with Math Reasoning and Chart Derendering},
  author={Liu, Fangyu and Piccinno, Francesco and Krichene, Syrine and Pang, Chenxi and Lee, Kenton and Joshi, Mandar and Altun, Yasemin and Collier, Nigel and Eisenschlos, Julian},
  booktitle={Proceedings of the 61st Annual Meeting of the Association for Computational Linguistics (Volume 1: Long Papers)},
  pages={12756--12770},
  year={2023}
}

@inproceedings{lee2023pix2struct,
  title={Pix2struct: Screenshot parsing as pretraining for visual language understanding},
  author={Lee, Kenton and Joshi, Mandar and Turc, Iulia Raluca and Hu, Hexiang and Liu, Fangyu and Eisenschlos, Julian Martin and Khandelwal, Urvashi and Shaw, Peter and Chang, Ming-Wei and Toutanova, Kristina},
  booktitle={International Conference on Machine Learning},
  pages={18893--18912},
  year={2023},
  organization={PMLR}
}

@inproceedings{zhang2024improve,
    title = "Improve Vision Language Model Chain-of-thought Reasoning",
    author = "Zhang, Ruohong  and
      Zhang, Bowen  and
      Li, Yanghao  and
      Zhang, Haotian  and
      Sun, Zhiqing  and
      Gan, Zhe  and
      Yang, Yinfei  and
      Pang, Ruoming  and
      Yang, Yiming",
    editor = "Che, Wanxiang  and
      Nabende, Joyce  and
      Shutova, Ekaterina  and
      Pilehvar, Mohammad Taher",
    booktitle = "Proceedings of the 63rd Annual Meeting of the Association for Computational Linguistics (Volume 1: Long Papers)",
    month = jul,
    year = "2025",
    address = "Vienna, Austria",
    publisher = "Association for Computational Linguistics",
    url = "https://aclanthology.org/2025.acl-long.82/",
    doi = "10.18653/v1/2025.acl-long.82",
    pages = "1631--1662",
    ISBN = "979-8-89176-251-0",
}

@article{hegde2025chartqa,
  title={ChartQA-X: Generating Explanations for Charts},
  author={Hegde, Shamanthak and Fazli, Pooyan and Seifi, Hasti},
  journal={arXiv preprint arXiv:2504.13275},
  year={2025}
}

@article{ma2025sci,
  title={SCI-Reason: A Dataset with Chain-of-Thought Rationales for Complex Multimodal Reasoning in Academic Areas},
  author={Ma, Chenghao and Ding, Junpeng and Zhang, Jun and Ma, Ziyan and Qing, Huang and Gao, Bofei and Chen, Liang and Song, Meina and others},
  journal={arXiv preprint arXiv:2504.06637},
  year={2025}
}

@inproceedings{jiaqi2025think,
  title={Think or Not? Selective Reasoning via Reinforcement Learning for Vision-Language Models},
  author={Jiaqi, WANG and Lin, Kevin Qinghong and Cheng, James and Shou, Mike Zheng},
  booktitle={The Exploration in AI Today Workshop at ICML 2025},
  year={2025},
}

@inproceedings{meng2024chartassistant,
  title={ChartAssistant: A Universal Chart Multimodal Language Model via Chart-to-Table Pre-training and Multitask Instruction Tuning},
  author={Meng, Fanqing and Shao, Wenqi and Lu, Quanfeng and Gao, Peng and Zhang, Kaipeng and Qiao, Yu and Luo, Ping},
  booktitle={Findings of the Association for Computational Linguistics ACL 2024},
  pages={7775--7803},
  year={2024}
}

@inproceedings{zhang2024tinychart,
    title = "{T}iny{C}hart: Efficient Chart Understanding with Program-of-Thoughts Learning and Visual Token Merging",
    author = "Zhang, Liang  and
      Hu, Anwen  and
      Xu, Haiyang  and
      Yan, Ming  and
      Xu, Yichen  and
      Jin, Qin  and
      Zhang, Ji  and
      Huang, Fei",
    editor = "Al-Onaizan, Yaser  and
      Bansal, Mohit  and
      Chen, Yun-Nung",
    booktitle = "Proceedings of the 2024 Conference on Empirical Methods in Natural Language Processing",
    month = nov,
    year = "2024",
    address = "Miami, Florida, USA",
    publisher = "Association for Computational Linguistics",
    url = "https://aclanthology.org/2024.emnlp-main.112/",
    doi = "10.18653/v1/2024.emnlp-main.112",
    pages = "1882--1898",
}

@inproceedings{dai2025graph,
  title={Graph-Based Multimodal Contrastive Learning for Chart Question Answering},
  author={Dai, Yue and Han, Soyeon Caren and Liu, Wei},
  booktitle={Proceedings of the 48th International ACM SIGIR Conference on Research and Development in Information Retrieval},
  pages={2658--2663},
  year={2025}
}

@inproceedings{wu2024chartinsights,
    title = "{C}hart{I}nsights: Evaluating Multimodal Large Language Models for Low-Level Chart Question Answering",
    author = "Wu, Yifan  and
      Yan, Lutao  and
      Shen, Leixian  and
      Wang, Yunhai  and
      Tang, Nan  and
      Luo, Yuyu",
    editor = "Al-Onaizan, Yaser  and
      Bansal, Mohit  and
      Chen, Yun-Nung",
    booktitle = "Findings of the Association for Computational Linguistics: EMNLP 2024",
    month = nov,
    year = "2024",
    address = "Miami, Florida, USA",
    publisher = "Association for Computational Linguistics",
    url = "https://aclanthology.org/2024.findings-emnlp.710/",
    doi = "10.18653/v1/2024.findings-emnlp.710",
    pages = "12174--12200",
}

@article{lee2025evaluating,
  title={Evaluating step-by-step reasoning traces: A survey},
  author={Lee, Jinu and Hockenmaier, Julia},
  journal={arXiv preprint arXiv:2502.12289},
  year={2025}
}

@article{shao2024deepseekmath,
  title={Deepseekmath: Pushing the limits of mathematical reasoning in open language models},
  author={Shao, Zhihong and Wang, Peiyi and Zhu, Qihao and Xu, Runxin and Song, Junxiao and Bi, Xiao and Zhang, Haowei and Zhang, Mingchuan and Li, YK and Wu, Yang and others},
  journal={arXiv preprint arXiv:2402.03300},
  year={2024}
}

@article{turpin2023language,
  title={Language models don't always say what they think: Unfaithful explanations in chain-of-thought prompting},
  author={Turpin, Miles and Michael, Julian and Perez, Ethan and Bowman, Samuel},
  journal={Advances in Neural Information Processing Systems},
  volume={36},
  pages={74952--74965},
  year={2023}
}

@inproceedings{reimers-2019-sentence-bert,
  title = "Sentence-BERT: Sentence Embeddings using Siamese BERT-Networks",
  author = "Reimers, Nils and Gurevych, Iryna",
  booktitle = "Proceedings of the 2019 Conference on Empirical Methods in Natural Language Processing",
  month = "11",
  year = "2019",
  publisher = "Association for Computational Linguistics",
  url = "https://arxiv.org/abs/1908.10084",
}

@article{team2025gemma,
  title={Gemma 3 technical report},
  author={Team, Gemma and Kamath, Aishwarya and Ferret, Johan and Pathak, Shreya and Vieillard, Nino and Merhej, Ramona and Perrin, Sarah and Matejovicova, Tatiana and Ram{\'e}, Alexandre and Rivi{\`e}re, Morgane and others},
  journal={arXiv preprint arXiv:2503.19786},
  year={2025}
}

@article{wang2025internvl3,
  title={Internvl3. 5: Advancing open-source multimodal models in versatility, reasoning, and efficiency},
  author={Wang, Weiyun and Gao, Zhangwei and Gu, Lixin and Pu, Hengjun and Cui, Long and Wei, Xingguang and Liu, Zhaoyang and Jing, Linglin and Ye, Shenglong and Shao, Jie and others},
  journal={arXiv preprint arXiv:2508.18265},
  year={2025}
}

@article{liu2025more,
  title={More Thinking, Less Seeing? Assessing Amplified Hallucination in Multimodal Reasoning Models},
  author={Liu, Chengzhi and Xu, Zhongxing and Wei, Qingyue and Wu, Juncheng and Zou, James and Wang, Xin Eric and Zhou, Yuyin and Liu, Sheng},
  journal={arXiv preprint arXiv:2505.21523},
  year={2025}
}

@article{chen2024expanding,
  title={Expanding performance boundaries of open-source multimodal models with model, data, and test-time scaling},
  author={Chen, Zhe and Wang, Weiyun and Cao, Yue and Liu, Yangzhou and Gao, Zhangwei and Cui, Erfei and Zhu, Jinguo and Ye, Shenglong and Tian, Hao and Liu, Zhaoyang and others},
  journal={arXiv preprint arXiv:2412.05271},
  year={2024}
}

@article{schulman2017proximal,
  title={Proximal policy optimization algorithms},
  author={Schulman, John and Wolski, Filip and Dhariwal, Prafulla and Radford, Alec and Klimov, Oleg},
  journal={arXiv preprint arXiv:1707.06347},
  year={2017}
}

@article{masrybigcharts,
  title={Bigcharts-r1: Enhanced chart reasoning with visual reinforcement finetuning},
  author={Masry, Ahmed and Puri, Abhay and Hashemi, Masoud and Rodriguez, Juan A and Thakkar, Megh and Mahajan, Khyati and Yadav, Vikas and Madhusudhan, Sathwik Tejaswi and Pich{\'e}, Alexandre and Bahdanau, Dzmitry and others},
  journal={arXiv preprint arXiv:2508.09804},
  year={2025}
}

@article{wang2024charxiv,
  title={Charxiv: Charting gaps in realistic chart understanding in multimodal llms},
  author={Wang, Zirui and Xia, Mengzhou and He, Luxi and Chen, Howard and Liu, Yitao and Zhu, Richard and Liang, Kaiqu and Wu, Xindi and Liu, Haotian and Malladi, Sadhika and others},
  journal={Advances in Neural Information Processing Systems},
  volume={37},
  pages={113569--113697},
  year={2024}
}

@inproceedings{deyoung2020eraser,
  title={ERASER: A benchmark to evaluate rationalized NLP models},
  author={DeYoung, Jay and Jain, Sarthak and Rajani, Nazneen Fatema and Lehman, Eric and Xiong, Caiming and Socher, Richard and Wallace, Byron C},
  booktitle={Proceedings of the 58th annual meeting of the association for computational linguistics},
  pages={4443--4458},
  year={2020}
}

@inproceedings{tutek2025measuring,
  title={Measuring chain of thought faithfulness by unlearning reasoning steps},
  author={Tutek, Martin and Chaleshtori, Fateme Hashemi and Marasovi{\'c}, Ana and Belinkov, Yonatan},
  booktitle={Proceedings of the 2025 Conference on Empirical Methods in Natural Language Processing},
  pages={9946--9971},
  year={2025}
}

@article{paul2024making,
  title={Making reasoning matter: Measuring and improving faithfulness of chain-of-thought reasoning},
  author={Paul, Debjit and West, Robert and Bosselut, Antoine and Faltings, Boi},
  journal={arXiv preprint arXiv:2402.13950},
  year={2024}
}

@article{guan2025monitoring,
  title={Monitoring monitorability},
  author={Guan, Melody Y and Wang, Miles and Carroll, Micah and Dou, Zehao and Wei, Annie Y and Williams, Marcus and Arnav, Benjamin and Huizinga, Joost and Kivlichan, Ian and Glaese, Mia and others},
  journal={arXiv preprint arXiv:2512.18311},
  year={2025}
}

@inproceedings{Wei2022CoT,
  title        = {Chain-of-Thought Prompting Elicits Reasoning in Large Language Models},
  author       = {Wei, Jason and Wang, Xuezhi and Schuurmans, Dale and Bosma, Maarten and Ichter, Brian and Xia, Fei and Chi, Ed and Le, Quoc and Zhou, Denny},
 booktitle = {Advances in Neural Information Processing Systems},
 editor = {S. Koyejo and S. Mohamed and A. Agarwal and D. Belgrave and K. Cho and A. Oh},
 pages = {24824--24837},
 publisher = {Curran Associates, Inc.},
 url = {https://proceedings.neurips.cc/paper_files/paper/2022/file/9d5609613524ecf4f15af0f7b31abca4-Paper-Conference.pdf},
 volume = {35},
 year = {2022}
}

@article{sinha2026attention,
  title={Attention-guided Fine-tuning of Multimodal Large Language Models Improves Chain-of-Thought Reasoning},
  author={Sinha, Sanchit and Xiong, Guangzhi and Liu, Bohan and He, Zhenghao and Zhang, Aidong},
  journal={arXiv preprint arXiv:2606.01558},
  year={2026}
}

\appendix

\newpage

\clearpage

\appendix
\section{Appendix}  
\label{sec:appendix}

\subsection{Salient Dataset Construction Details}

\noindent\textbf{Chart-RVR CoT Reasoning Datasets.} Although the training datasets discussed contain plenty of examples, there is a distinct lack of a reliable source of ground-truth rationales, data tables, and chart-type annotations associated with them. As a consequence, we generate a CoT Chart dataset sampled from the training splits of ChartQA, ChartFC, and PlotQA. To generate faithful CoT rationales, inferring the chart type and generating the associated data tables, we utilize a large-scale SOTA LVLM - Qwen2.5VL-72B. The prompt template for generating the dataset is provided in Figure~\ref{fig:dset-gen-prompt} (Appendix), where both the query and label are provided to the model. (1) \textit{CoT Datasets:} For the CoT dataset, we randomly sample 2,000 datapoints each from the aforementioned datasets based on a specific seed for a total of 6,000 training samples. (2) \textit{CoT-Hard Dataset}: A significant issue in randomly sampling training points from the datasets is the lack of diversity and dominance of easy samples, which constitute queries that require no reasoning, e.g., `title of the chart. To alleviate this, we specifically sample data from the human-annotated reasoning subset of ChartQA (labeled `human') and random samples from the ChartFC and PlotQA data. We filter out overly simplistic questions from the dataset for a total of 30,000 training samples. 

\noindent \textbf{Chart Types ($\mathcal{C}$).}
Even though a large variety of chart types exist, most of the commonly found charts can be categorized into 10 fundamental categories. We instantiate a set of chart families: \emph{Line}, \emph{Bar}, \emph{Stacked Bar}, \emph{Pie}, \emph{Histogram}, \emph{Scatter}, \emph{Area}, \emph{Stacked Area}, \emph{Bubble}, and \emph{Treemap}.
\footnote{We normalize synonyms to a canonical label: \textit{Column}$\rightarrow$\textit{Bar}, \textit{Donut}$\rightarrow$\textit{Pie}, \textit{Point}$\rightarrow$\textit{Scatter}. Grouped bars are labeled \textit{Bar}; cumulative variants are \textit{Stacked Bar}.}

\noindent \textbf{Construction.} To generate the Chart-RVR CoT datasets, we utilize a large open-source LVLM as an oracle, namely Qwen2.5VL-72B-Instruct \citep{qwen2.5-VL}. For the Chart-RVR datasets, the prompt template utilized is shown in Figure~\ref{fig:dset-gen-prompt}. 

\begin{enumerate}[parsep=0pt, itemsep=0pt, topsep=0pt]
    \item \textit{Normalization.} Images are resized with aspect-ratio preservation to a minimum of 512 leading edge size; tables are canonicalized to a header (columns) + rectangular body (rows) with string conversion where applicable. We map the dataset's native chart label to the chart types above.
    \item \textit{Templated prompting.} We instantiate Figure~\ref{fig:dset-gen-prompt} with (a) the canonical chart label, (b) a compact task description, and (c) formatting constraints (typed tags, JSON schema, operation tokens). The template explicitly separates (i) \emph{data gathering} steps from (ii) \emph{computation} steps to enable process supervision.
    \item \textit{Sampling and pre-screen.} For each (image, question) pair, we output rationales via deterministic inference with temperature set to 0. Candidates failing structural checks (JSON parse, length bounds, required tags) are discarded before reward computation.
\end{enumerate}
\textbf{NOTE: As opposed to many other approaches, our dataset can be constructed from pre-existing datasets, imparting transparency.}

\noindent \textbf{Manual Filtering of sub-par rationales.} Although our rewards prune many low-quality generations automatically, we perform a light manual pass to remove pathological rationales that would otherwise pollute training data points. A rationale is marked \textit{sub-par} and discarded if any of the following holds:
\begin{enumerate}[parsep=0pt, itemsep=0pt, topsep=0pt]
    \item \textit{Unparseable structure:} the emitted JSON blocks (table tag) fail to parse or violate schema (missing keys, non-rectangular rows).
    \item \textit{Data hallucination:} The rationale cites a category/series not present in the ground-truth table, i.e., non-sensical characters, special symbols, etc.
    \item \textit{Length:} We filter out all rationales with fewer than 3 steps of reasoning.
\end{enumerate}

We utilize the following procedure for final dataset validation:
\begin{enumerate}
    \item For pass 1, the prompt structure shown in Figure~\ref{fig:dset-gen-prompt} is used to generate the CoT data. 
    \item For pass 2, all samples with overtly short/long and incorrect reasoning traces are filtered (i.e. trace length less than 3 or more than 8 based on manual inspection and outlier rejection). To validate if a trace is wrong, we algorithmically check if the last CoT line trace contains the correct answer. A total of 358 samples are filtered out in this step out of 15000. 
    \item For pass 3, three doctoral-level researchers are employed to validate the reasoning traces of a randomly sampled data subset (1000 samples). Each researcher filtered fewer than 6 samples (1\%) in total, with a 100\% agreement between them.
\end{enumerate}

We will release the filtered dataset upon acceptance for transparency and faithful reproduction. Additionally, the smaller CoT Dataset with appropriate seeds will also be released.

\noindent \textbf{Agreement of annotation done using human validation}

\begin{figure}[h]
    \centering
    \includegraphics[width=\linewidth]{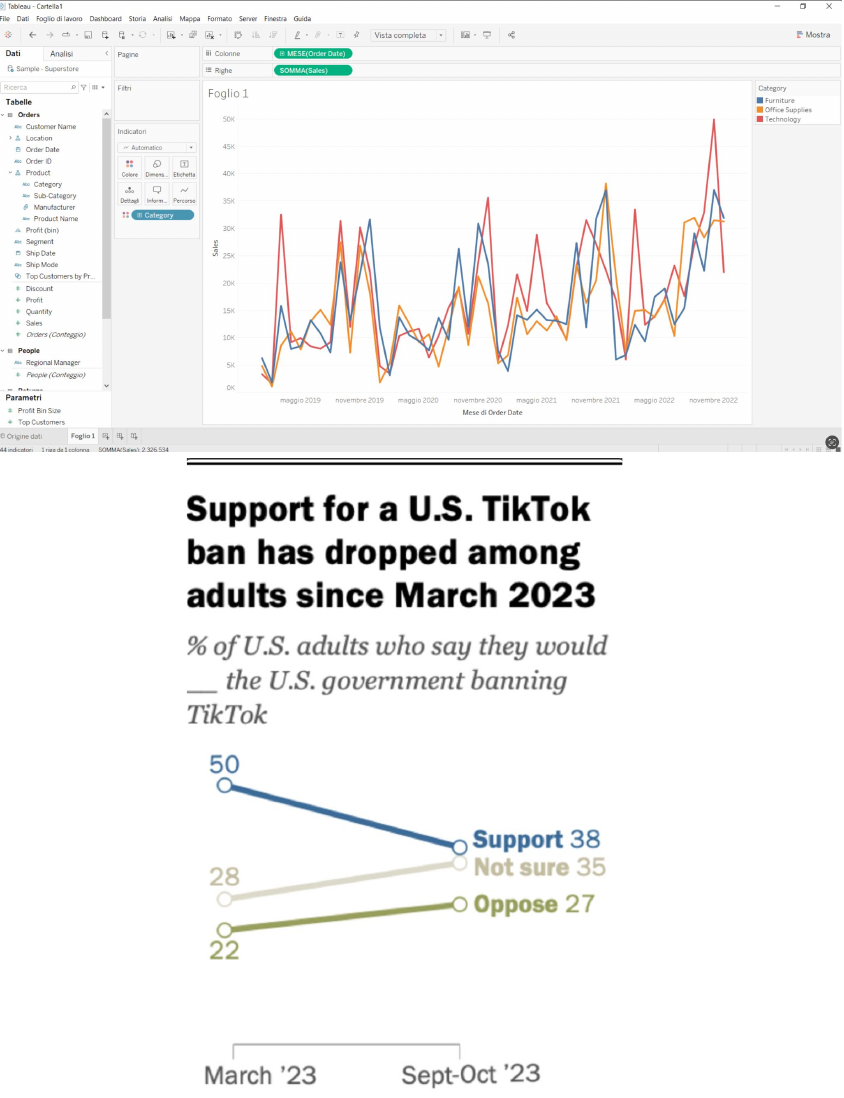}
    \caption{Examples of charts where the Oracle and human annotations (TOP) \textbf{need approximation} and (BOTTOM) are \textbf{factoid/direct}; the corresponding annotated data tables may differ.}
    \label{fig:failure-case-annotation}
\end{figure}

We manually validated 100 randomly sampled oracle-annotated charts, each from the EvoChart and ChartQAPro datasets (note no ground truth table annotations are available for either). We categorize each data sample into 2 categories: (1) Needs Approximation? - which entails guessing the correct values from the chart axes rather than labelled counterparts, "Factoid/Direct" - which entails looking at axes marks to determine the correct values. In Table~\ref{tab:human-oracle-kappa}, we report the Cohen's $\kappa$: the inter-annotator agreement score b/w human (Annotator-1) and Oracle model (Annotator-2). We also show examples of the categorization in Figure~\ref{fig:failure-case-annotation}.

\begin{table}[h]
\centering
\begin{tabular}{lcc}
\toprule
Category            & \#Samples & Cohen's $\kappa$ \\
\midrule
Needs Approximation & 48        & 0.91             \\
Factoid/Direct      & 152       & 0.98             \\
\textbf{Overall}    & 200       & 0.97             \\
\bottomrule
\end{tabular}
\caption{Inter-annotator agreement between human annotators and the Oracle model.}
\label{tab:human-oracle-kappa}
\end{table}

\begin{figure}[h]
    \centering
    \includegraphics[width=\linewidth]{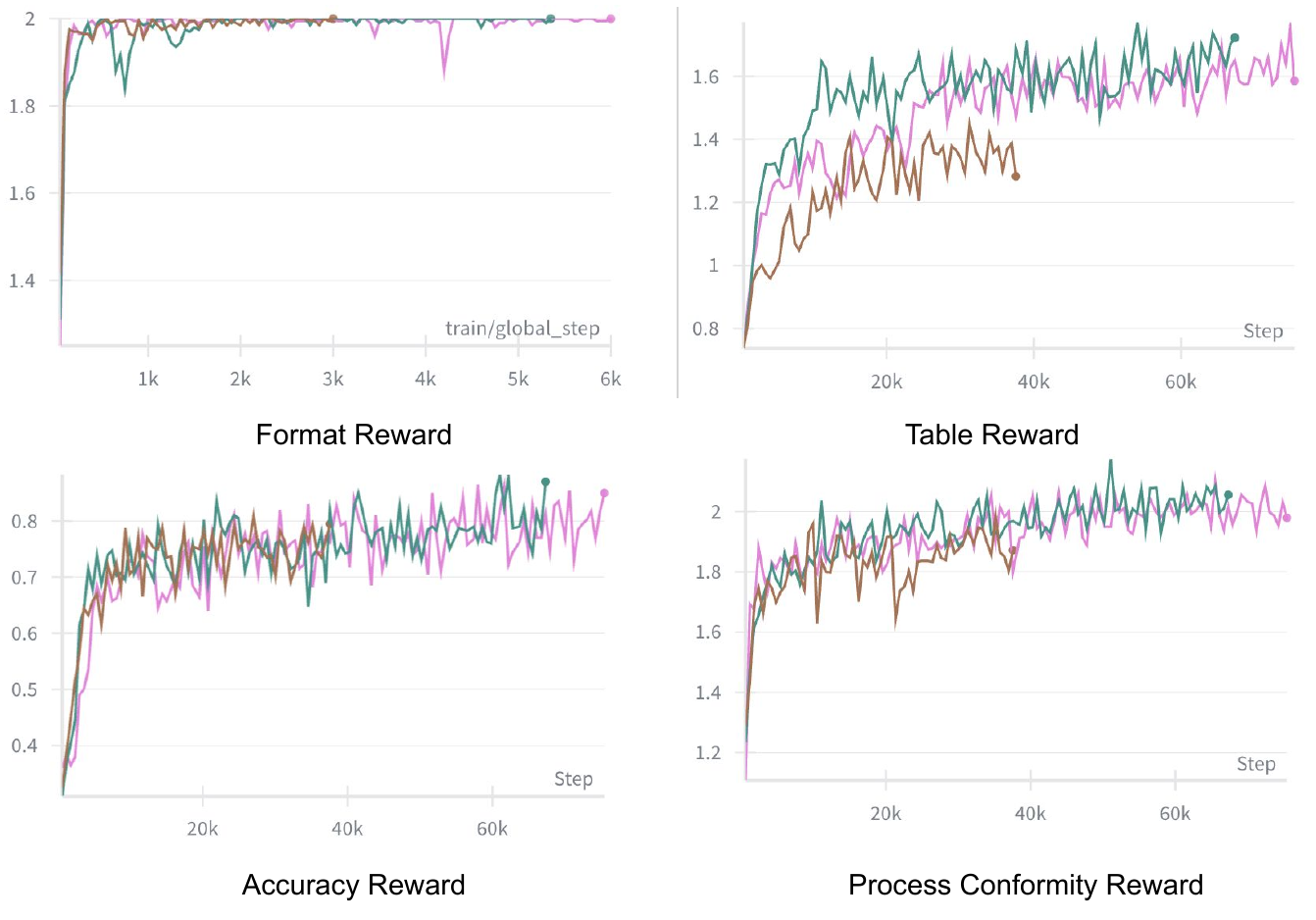}
    \caption{Chart-RVR Reward maximization during training on 3 separate CoT datasets on Qwen2.5VL-3B.}
    \label{fig:training-curve}
\end{figure}

\begin{figure}[!htbp]
\centering
\begin{tcolorbox}[colback=grey,colframe=black,title=System Prompt template for CoT Dataset Generation]
\small
You are helping me answer questions on charts. \\
You have to look at chart picture and question. \\
The question and the answer will be provided to you. \\
First you have to recover the table data from the chart image in JSON format.\\
For the chart image, output only a JSON object with: \\
"columns": list of column headers, \\
"rows": list-of-lists, one per data row \\
No prose, no comments. \\
1. Respond with **only** a JSON object inside a ```json code fence. \\
2. The JSON must use exactly this schema: \\
    \{\\
        "columns": [...],\\
        "rows": [...]\\
    \} \\
3. Do NOT output HTML, Markdown, or commentary. Any deviation gets zero reward.\\
Next, think step by step in as many small steps as required to answer the question based on the chart.\\
Lastly, also predict the type of chart out of the following:\\
"line", "bar", "stacked bar", "pie", "histogram", "scatterplot", "area", "stacked area", "bubble", "treemap" \\         
Format:\\
\#\#\# Question: $<$question$>$ \\
\#\#\# Answer: $<$answer$>$ \\
\#\#\# Table: $<$json table$>$ \\
\#\#\# Reasoning:  \\
$<$step-1$>$: Provide a description of reasoning \\
$<$step-2$>$: Gather ALL the appropriate data from the chart \\
$<$step-3$>$: Break down the query into smaller parts and verify each part with the data \\
... \\
$<$step-n$>$: Do the final calculation or reasoning to derive the answer \\
$<$step-n+1$>$: VERIFY the final answer is correct for no hallucinations \\ 
\#\#\# Type: $<$type of chart$>$ \\
\end{tcolorbox}
\caption{\textbf{System prompt template} used across Structured CoT, SFT, GRPO, and Chart-RVR.}
\label{fig:dset-gen-prompt}
% \vspace{-10pt}
\end{figure}

\noindent \textbf{Emphasizing OOD benchmarks.} Many widely used chart reasoning datasets like ChartQA, PlotQA, and ChartFC have been repeatedly incorporated (in whole or in part) into instruction-tuning corpora, synthetic data expansions, and public multimodal mixtures that contemporary LVLMs are trained or aligned on, together with related datasets such as FigQA and DVQA\footnote{Note: Because most LVLM training mixtures are only partially disclosed, we cannot exhaustively audit every provider. Our designation reflects (i) the public availability and long shelf-life of ChartQA/PlotQA/ChartFC, (ii) their documented use in multiple open instruction-tuning recipes as documented in InternVL's technical report \citep{chen2024expanding}}. This creates a realistic risk of distributional familiarity and data leakage at pretraining/finetuning time, yielding optimistic ``in-distribution'' estimates. To assess generalization beyond such exposure, we therefore evaluate on `truly' out-of-distribution (OOD) results on EvoChart, ChartBench, and ChartQA-Pro, which were not used in model pretraining or alignment and contain chart styles, templates, and question programs that differ from the legacy sets. As shown in the Appendix, it can be observed that datasets like EvoChart possess significantly harder chart images than benchmark sets like ChartQA. Throughout the paper, we treat ChartQA/PlotQA/ChartFC as `ID' baselines and use EvoChart/ChartBench/ChartQA-Pro to measure robustness, reasoning fidelity, and explainability under genuine distribution shift. Truly generalizable chart reasoning with explanations remains an open problem for frontier LVLMs.

\subsection{Implementation Details}

\noindent\textbf{Training Details.} \textbf{SFT:} We utilize the same system prompt format as in Figure~\ref{fig:grpo-prompt} for fine-tuning. We train the entire model for 3 epochs, with a learning rate of $1e-5$ for the LLM and projector, while $2e-6$ for the vision tower, with a warm-up ratio of $0.03$. \textbf{Chart-RVR:} We utilized the TRL \citep{vonwerra2022trl} implementation of GRPO with a maximum prompt length of 4096, maximum completion length 768, and number of generations (rollouts) 4 per sample. To further reduce prompt lengths, we utilize the JSON notation to represent the underlying chart tables. For Process Conformity Rewards, we used sentence embeddings from a lightweight embedding model, MiniLM-L6-v2 \citep{reimers-2019-sentence-bert}. We train all models for 4 epochs with a learning rate set as $1e-6$.

\noindent For SFT, each training sample's CoT trace and answer is appropriately formatted into the expected prompt format and adhering to \texttt{<think><type></type><table></table>\\...</think><answer></answer>}. The validation dataset is created using \textbf{500} randomly sampled data points from the training sets of ChartQA, ChartFC, and PlotQA. 
\begin{itemize}
    \item \textbf{SFT:} For SFT, we train the model for a maximum of 3 epochs and select the model with the highest accuracy on the validation set. Additionally, we utilize the AdamW optimizer for training with a training batch size of 4 across 2 NVIDIA H100 GPUs. The learning rate schedule utilized is linear with a maximum learning rate of $1e-5$ and 1000 warmup steps. During training, the entire model is trained with FP16 precision, and the vision tower is trained using an LR of $2e-6$.
    \item \textbf{Chart-RVR:} For Qwen2.5VL models, we utilize a maximum completion length of 768 and a total prompt length of 4096. For Gemma and InternVL models, due to more efficient vision token computations, we utilize a maximum prompt length of 3072 with a maximum completion length of 512. For Qwen2.5VL, we utilize a maximum learning rate of $1e-6$ while for Gemma and InternVL, the learning rate is reduced to $5e-7$ and $2e-7$ respectively. The training process takes place on 4 NVIDIA H100 GPUs with a per-device batch size of 2. The number of rollouts is set to 4. The total number of epochs is set at 4 for all models and configurations.
\end{itemize}

\noindent \textbf{Hyperparameter Tuning.} Note that the Chart-RVR setup has 3 major hyperparameters to tune, which assign relative weights to each component of the reward design. Unless otherwise stated, we use fixed values across all model families: $\lambda_1=1$ for the surrogate-task reward and $\lambda_2=1$ for the process-conformity reward.
In practice, we found the method to be insensitive to modest perturbations of $\lambda_1$, $\lambda_2$, and the compute budget (within $\pm 1$ for $\lambda_1,\lambda_2$), due to the smooth normalization of the combined reward. With sufficient optimization steps, each reward term saturates, and the final ranking of methods remains unchanged. We therefore fix $\lambda_1$ and $\lambda_2$ to simplify training and ensure comparability across backbones.

\noindent \textbf{Evaluation Setup.} For all our evaluations, we utilize the public test sets of the datasets. For ChartQA\footnote{\url{https://huggingface.co/datasets/HuggingFaceM4/ChartQA}}, ChartQAPro\footnote{\url{https://huggingface.co/datasets/ahmed-masry/ChartQAPro}}, EvoChart\footnote{\url{https://github.com/MuyeHuang/EvoChart}}, and ChartFC\footnote{\url{https://github.com/mubasharaak/ChartCheck}}, we utilize the entire test sets sampled from sources as listed in the footnotes. For ChartBench\footnote{\url{https://huggingface.co/datasets/SincereX/ChartBench}} and PlotQA\footnote{\url{https://github.com/NiteshMethani/PlotQA/blob/master/PlotQA_Dataset.md}}, due to their massive sizes, we sample a 5000-sized subset using a fixed seed. For all models, we resize the leading image edges to a minimum and maximum of 448 and 812, respectively, using Bicubic interpolation. For all structured CoT, SFT, and GRPO evaluations, we utilize the prompt template in Figure~\ref{fig:grpo-prompt} as the system prompt for all models.

\begin{figure}[!htbp]
\centering
% \scriptsize
\begin{tcolorbox}[colback=grey,colframe=black,title=System Prompt template for Structured CoT / SFT / GRPO / Chart-RVR]
\small
You are a vision-language assistant. You are given a chart image and a query about the chart. 

Think step-by-step about how to answer the query based on the chart image and then provide the final answer.

$\#\#\#$ Output format:

Respond **with exactly two blocks in order and nothing else**:

$<$think$>$

First output the type of chart in $<$type$>$, \
then output the underlying data table and finally, \ 
think step-by-step about how to answer the query based on the chart image \
and then provide the final answer.
$<$type$>$
Type of chart - one word from line, bar, stacked bar, pie, histogram, scatterplot, area, stacked area, bubble, treemap.
$<$/type$>$

Next output the data table in the $<$table$>$$<$/table$>$ tags

$<$table$>$

json table - for the chart image, output only a JSON object with: "columns": list of column headers, "rows": list-of-lists, one per data row

No prose, no comments.

1. Respond with **only** a JSON object\\
2. The JSON must use exactly this schema:
    \{
        "columns": [...],
        "rows": [[...], [...],..., [...]]
    \}\\
3. Do NOT output HTML, Markdown, or commentary. Any deviation gets zero reward.

$<$/table$>$

Provide your reasoning here in steps:\\
$<$step-1$>$: Provide a description of reasoning\\
$<$step-2$>$: Gather ALL the appropriate data from the chart\\
$<$step-3$>$: Break down the query into smaller parts and verify each part with the data\\
...\\
$<$step-n$>$: Do the final calculation or reasoning to derive the answer\\
$<$/think$>$\\
$<$answer$>$\\
Final answer on a single line\\
$<$/answer$>$
\end{tcolorbox}
\caption{\textbf{System prompt template} used across Structured CoT, SFT, GRPO, and Chart-RVR.}
\label{fig:grpo-prompt}
\normalsize
\end{figure}

\subsection{Practical Reward Hacking and Mitigation Strategies}
Reward hacking is a common phenomenon in Reinforcement Fine-Tuning (RFT), 
In this section, we demonstrate common reward-hacking behavior observed in our experiments and mitigation strategies employed by us to alleviate these concerns. Note that the reward hacking behavior is usually observed during the early stages of the training and can cause a sudden training collapse, from which the policy never recovers.

\noindent \textbf{Stacked Rewards on Length.} As the models are incentivized to output longer traces (rollouts) via the length reward, in some cases, this results in unrelated characters being outputted to `fill' in the extra tokens by new line tokens (`\textbackslash n') or repeating multiple identical redundant steps. This is often observed when smaller LVLMs, which do not output their CoT traces reliably, are suddenly incentivized to output much longer traces. If left unmitigated, this can cause a training collapse where the model can never recover from this particular policy, which technically still maximizes the reward. To alleviate this, we utilize a stacked reward design, where a partial reward is assigned for reasoning lengths above certain thresholds. This treats the length reward maximization as a `warm' start process and nudges the model to output longer traces gradually and not immediately upon the start of training. We utilize a 0.5 reward for token lengths of 100 or more, and finally, the full 1.0 is set when the length exceeds $\eta_1$ tokens. Additionally, we also penalize `filler' tokens by checking if more than 5 `\textbackslash n' occur in contiguous tokens; we provide a 0 reward.

\noindent \textbf{Stacked Rewards on Table Reconstruction.}
As detailed in the main text, we utilize the Table reconstruction rewards in tranches -  a successful JSON parse of the tokens inside each rollout's $<$table$>$ tags gives a 0.5 reward. As the reward function on table reconstruction is extremely dense, we warm start the process by assigning a 0.5 extra reward if `columns' and `rows' appear in the rollout's JSON parsing.

\subsection{Training Curves and Reward Maximization Behavior}
In Figure~\ref{fig:training-curve}, we visually demonstrate the reward maximization during the Chart-RVR training process on 3 splits of the CoT dataset. We observe that the format rewards are maximized very early in the training with minimal changes observed throughout the process. The accuracy reward gradually improves and stagnates after about 60k steps (about 2 epochs). Similarly, the Table Rewards and Process Conformity Rewards demonstrate a smooth increase initially and then are maximized around the same number of steps. All rewards show a massive improvement during early training.

\begin{table}[h]
\centering
\caption{Comparison between fine-tuned models using 3 distinct seeds of the ChartRVR-CoT Dataset. Chart-RVR improves performance over SFT across all datasets, with OOD improvements more pronounced.}
\resizebox{\linewidth}{!}{
\begin{tabular}{c|c|c|c|c}
\toprule
\textbf{Dataset} & \textbf{Direct} & \textbf{CoT} & \textbf{SFT} & \textbf{Chart-RVR} \\
\midrule
 ChartQA & 82.0 & 73.12 & 83.18 $\pm$ 0.33 & \textbf{83.87 $\pm$ 0.68} \\
 PlotQA & \textbf{80.50} & 52.72 & 76.05 $\pm$ 1.65& \textbf{78.71$\pm$0.20} \\
 ChartFC & 74.4 & 69.20 & 76.67 $\pm0.54$& \textbf{78.08$\pm$1.36}\\
 \midrule
 EvoChart & 48.72 & 29.6 & 46.50 $\pm$ 0.36 & \textbf{52.16$\pm$0.86} \\
 ChartQAPro & 25.7 & 15.80 & 24.52 $\pm$0.48 & \textbf{28.30$\pm$1.00}\\
 \bottomrule
\end{tabular}}
\label{tab:sft-vs-rvr}
\end{table}
\subsection{Consistent Improvement by Chart-RVR}
 We report the average performance and standard deviation in Table~\ref{tab:sft-vs-rvr} over all 3 seeds. As can be seen, Chart-RVR consistently outperformed SFT across both ID and OOD datasets by \textbf{1-2\%} and \textbf{4-6\%} respectively, demonstrating the efficacy of our method over SFT. In addition, the average results are at par with the results on a single CoT dataset split (as shown in Table~\ref{tab:chart_rvr_large}), highlighting robust convergence.

\subsection{Additional Explainability Results}
Table~\ref{tab:explanation-metric-appendix} reports the improvement on the explainability metric for LLaVA-Next-72B under three training modes (CoT, SFT, Chart-RVR). Chart-RVR delivers the largest gains on 4/5 datasets, notably +0.38 on ChartQA and +0.23 on EvoChart—while SFT narrowly leads on PlotQA (+0.1642 vs +0.1632). Overall gains are modest on harder sets (e.g., ChartQAPro = +0.058), but Chart-RVR provides the most consistent uplift in explainability.

\begin{table}[h]
\centering
\caption{Usefulness Results using LLaVA-Next-72B.}
\resizebox{0.9\linewidth}{!}{
\begin{tabular}{c|c|c|c}
\toprule
\textbf{Dataset} & \textbf{$\Delta$ CoT} & \textbf{$\Delta$ SFT} & \textbf{$\Delta$ Chart-RVR} \\
\midrule
 ChartQA & +0.27 & +0.29 & \bf +0.38 \\
 ChartQA (human) & +0.21 & +0.21 & \bf  +0.25 \\
 PlotQA & +0.1605 &  +0.1642 &  +0.1632 \\
 \midrule
 EvoChart & +0.16 & +0.14 & \bf +0.23  \\
 ChartQAPro & +0.049 & +0.043  & \bf +0.058 \\
 \bottomrule
\end{tabular}}
\label{tab:explanation-metric-appendix}
% \vspace{-5pt}
\end{table}

\subsection{LLM-as-Auditor Rubric}
\label{app:auditor}
We score two monitorability axes - Process Verifiability (PV) and Evidence Localization (EL) - with the system prompts below, issued to each oracle $\mathcal{O}\in\{\text{GPT-4o},\text{LLaVA-Next-72B}\}$. The chart image $x$, query $q$, and rationale $\hat a$ are supplied as the user turn; the oracle returns a single integer $1$--$5$ per axis. Identical prompts are used across oracles and conditions; scores are parsed by regex and averaged over the eval set.

% ---- shared box style ----
\tcbset{
  rubricbox/.style={
    colback=gray!4,
    colframe=black!70,
    boxrule=0.5pt,
    arc=2pt,
    left=6pt, right=6pt, top=5pt, bottom=5pt,
    fonttitle=\bfseries\small,
    fontupper=\footnotesize\ttfamily,
    breakable,
  }
}

% ============================================================
% PROCESS VERIFIABILITY
% ============================================================
\begin{tcolorbox}[rubricbox, title={System Prompt - Process Verifiability (PV)}]
You are a strict auditor of chart-reasoning rationales. You are given a chart
image, a question, and a model's step-by-step rationale. Judge ONLY
PROCESS VERIFIABILITY: whether the rationale is decomposed into discrete,
independently checkable steps that explicitly state the intermediate
quantities and the operations applied to them.

Do NOT reward fluency, length, or confidence. Do NOT judge whether the final
answer is correct. A long, well-written rationale with no stated numbers or
unseparated steps must score LOW.

Score on this 1-5 scale:
5 - Every step is discrete and self-checkable; each cited value and each
    operation (read, compare, add, average, etc.) is explicitly stated, so a
    reader could re-verify the whole chain without guessing.
4 - Mostly discrete and checkable; one step omits a value or merges two
    operations, but the chain is still auditable.
3 - Partially structured; some steps state quantities/operations, others are
    vague or bundle several actions into one.
2 - Largely unstructured; quantities or operations are mostly implicit and the
    chain cannot be checked step by step.
1 - A single block or free-form paragraph with no separable steps or stated
    intermediate quantities.

Output format: return ONLY the integer score, e.g. "4". No explanation.
\end{tcolorbox}

% ============================================================
% EVIDENCE LOCALIZATION
% ============================================================
\begin{tcolorbox}[rubricbox, title={System Prompt - Evidence Localization (EL)}]
You are a strict auditor of chart-reasoning rationales. You are given a chart
image, a question, and a model's step-by-step rationale. Judge ONLY
EVIDENCE LOCALIZATION: whether each reasoning step is grounded in a specific,
identifiable element of THIS chart (a named series, bar, slice, axis tick, or
table cell) rather than referring to the figure vaguely or citing values that
are not visible in it.

Use the chart image to verify grounding. Do NOT reward fluency or confidence.
Do NOT judge whether the final answer is correct. A step that asserts a number
with no identifiable source in the chart must score LOW, even if the number
happens to be right.

Score on this 1-5 scale:
5 - Every step names the specific chart element it reads from, and every cited
    value is traceable to a visible point/label in the chart.
4 - Nearly all steps are grounded to specific elements; at most one value is
    stated without a clear visual referent.
3 - Mixed; some steps point to specific elements while others reference the
    chart only generally ("the data shows...").
2 - Mostly ungrounded; few steps identify a specific element and several values
    have no visible source.
1 - No step is tied to an identifiable chart element; values appear without any
    traceable source.

Output format: return ONLY the integer score, e.g. "4". No explanation.
\end{tcolorbox}

% \newpage
\subsection{Visual Comparison between CoT, SFT, and Chart-RVR}
\begin{figure*}[h]
    \centering
    \includegraphics[width=\linewidth]{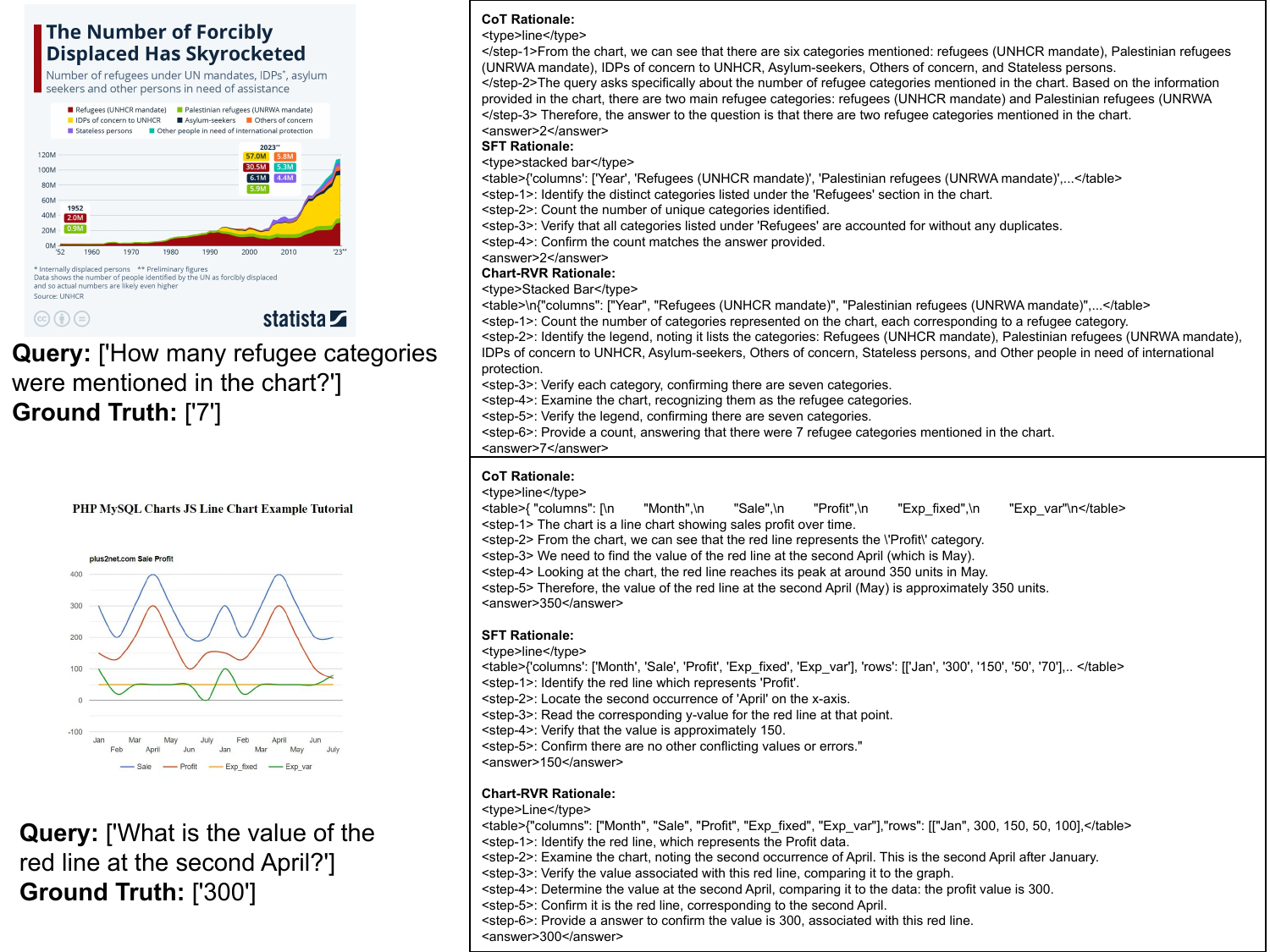}
    \caption{Example failure cases from the EvoChart dataset (OOD) where Chart-RVR outperforms CoT and SFT on 2 challenging examples.}
    \label{fig:failure-ood-1}
\end{figure*}

In this section, we demonstrate visual examples where Chart-RVR outperforms both CoT and SFT. First, we demonstrate Chart-RVR's effectiveness on OOD data on the EvoChart benchmark in Figure~\ref{fig:failure-ood-1}. As can be seen, CoT and SFT misidentify the number of categories as they only look at the legend on the left, while Chart-RVR correctly counts the number of colors. In the second case, Chart-RVR correctly identifies the value of the second April, which is a duplicate entry on the x-axis.

\begin{figure*}[h]
    \centering
    \includegraphics[width=\linewidth]{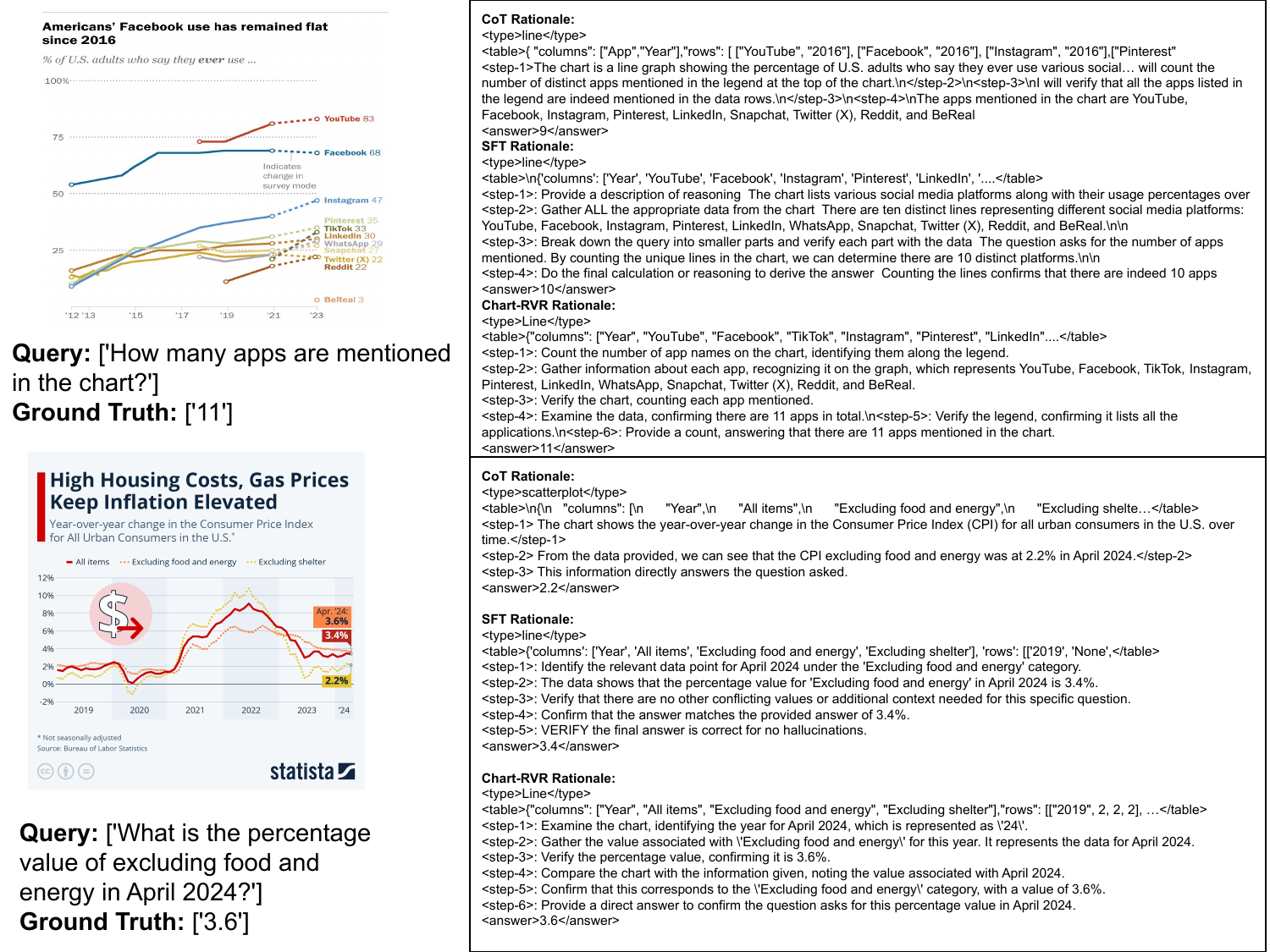}
    \caption{Example failure cases from the EvoChart dataset (OOD) where Chart-RVR outperforms CoT and SFT on 2 challenging examples.}
    \label{fig:failure-ood-2}
\end{figure*}

Next, in Figure~\ref{fig:failure-ood-2}, Chart-RVR correctly identifies the `BeReal' category, which is not picked up by both CoT and SFT, as it is only one data point on the line chart. This reinforces the need to condition reasoning on faithful table reconstruction. In the next image, Chart-RVR is able to capture the entity `3.6' accurately, where both CoT and SFT fail.

\begin{figure*}[h]
    \centering
    \includegraphics[width=\linewidth]{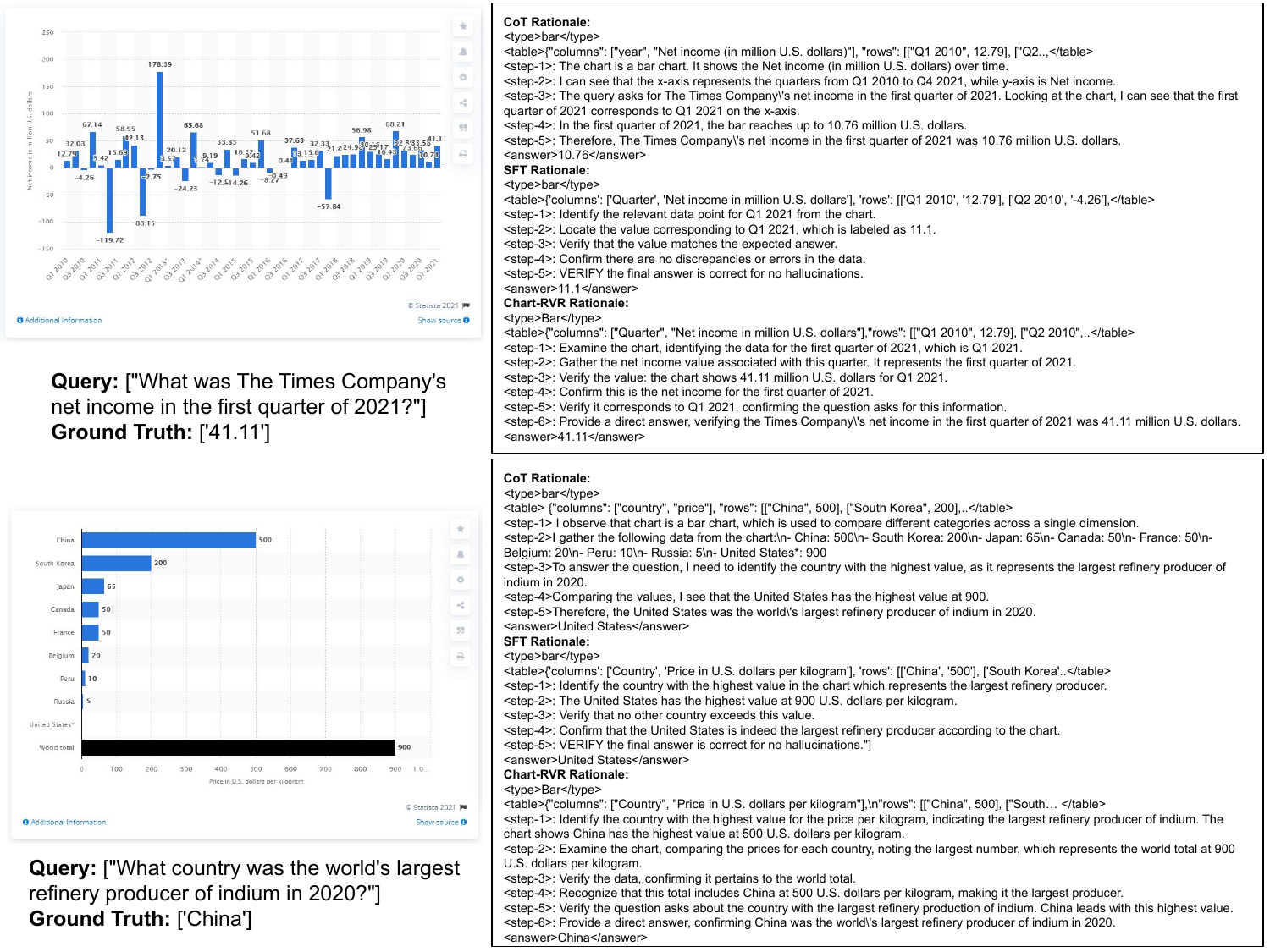}
    \caption{Example failure cases from the ChartQA dataset (ID) where Chart-RVR outperforms CoT and SFT on 2 challenging examples.}
    \label{fig:id-1}
\end{figure*}

\begin{figure*}[h]
    \centering
    \includegraphics[width=\linewidth]{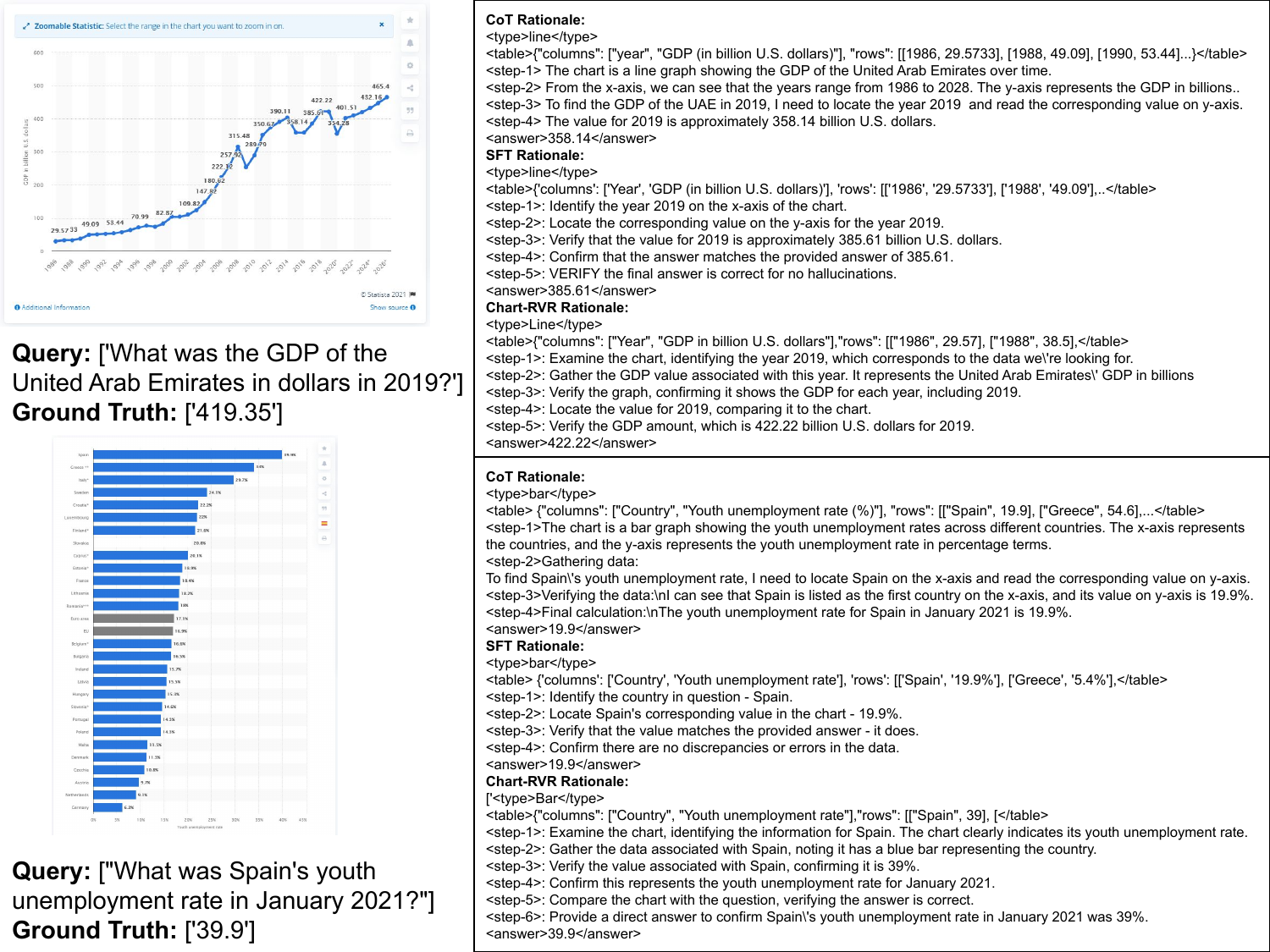}
    \caption{Example failure cases from the ChartQA dataset (ID) where Chart-RVR outperforms CoT and SFT on 2 challenging examples.}
    \label{fig:id-2}
\end{figure*}

\begin{figure*}[h]
    \centering
    \includegraphics[width=\linewidth]{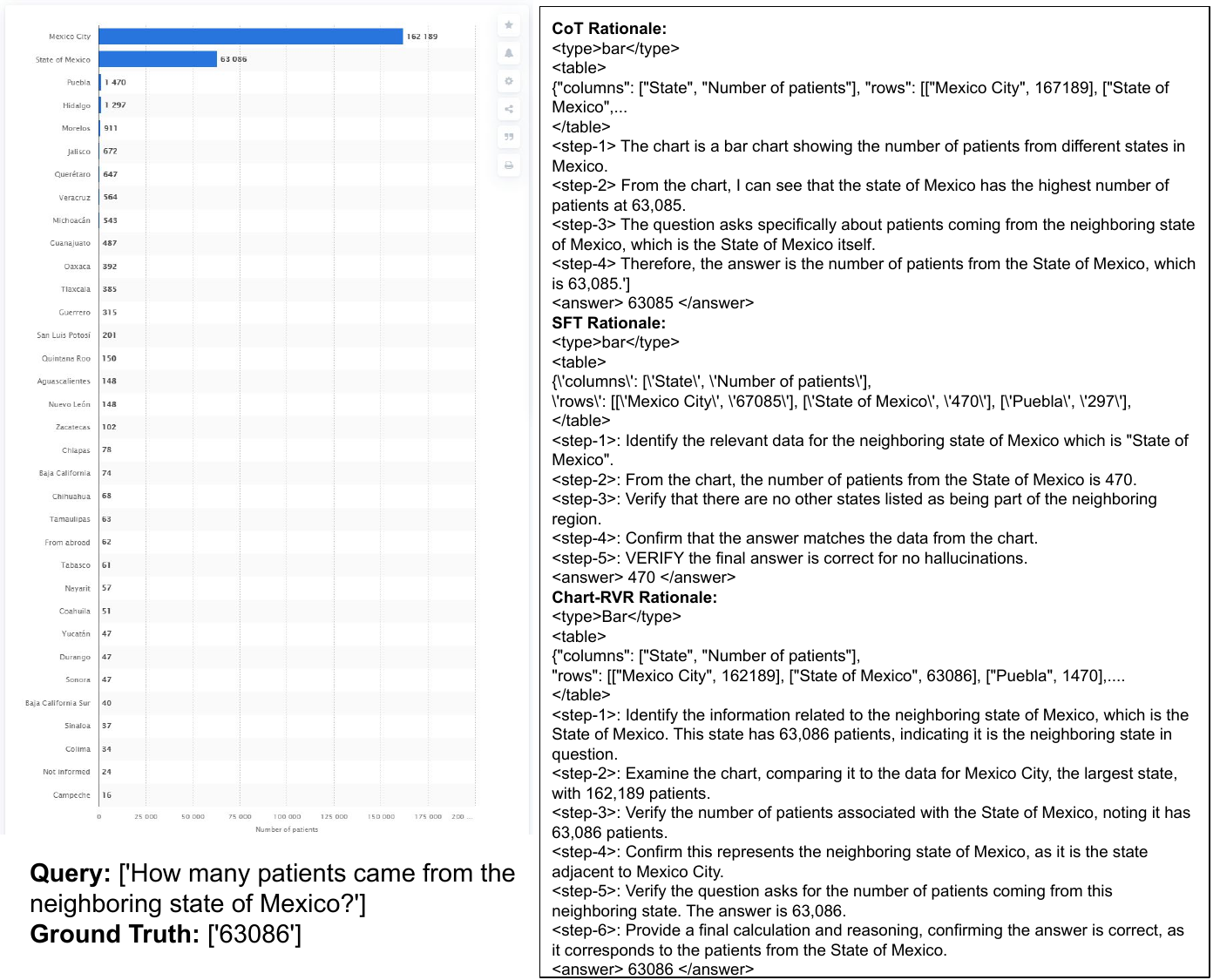}
    \caption{Example failure case from the ChartQA dataset (ID) where Chart-RVR outperforms CoT and SFT}
    \label{fig:id-3}
\end{figure*}

In Figures~\ref{fig:id-1},\ref{fig:id-2}, and \ref{fig:id-3}, we report cases where Chart-RVR outperforms both CoT and SFT on the ChartQA dataset. We observe that Chart-RVR is particularly accurate in cases where the chart is extremely complex.

% \newpage

\subsection{Failure Cases}
\begin{figure*}[h]
    \centering
    \includegraphics[width=\linewidth]{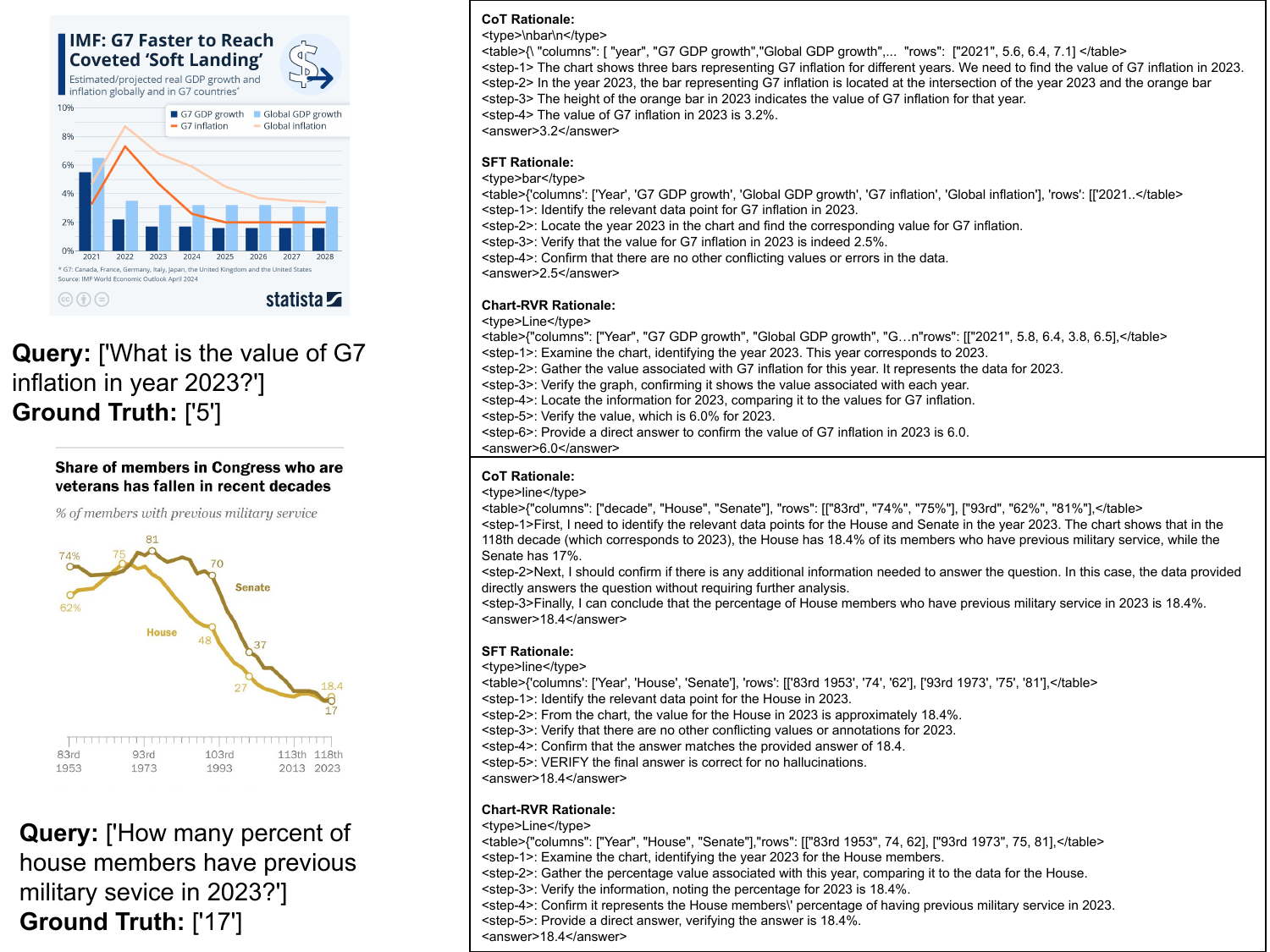}
    \caption{Example failure case from the EvoChart dataset (OOD) where all Chart-RVR, CoT, and SFT output wrong answers.}
    \label{fig:failure-cases}
\end{figure*}

Finally, in Figure~\ref{fig:failure-cases}, we report 2 failure cases of Chart-RVR on the OOD dataset EvoChart. The top chart is composed of both line and bar graphs together, making the reasoning process confounded. Note that Chart-RVR correctly identifies the chart type, i.e., line, which is the correct chart to look for, but answers the approximate value 6, which is very close to 5 and can be attributed to a scaling issue. In the next chart, the values of both lines in 2023 are extremely overlapping; hence, all approaches make errors.


% \newpage
\subsection{Human Study}
\begin{figure*}[h]
    \centering
    \includegraphics[width=\linewidth]{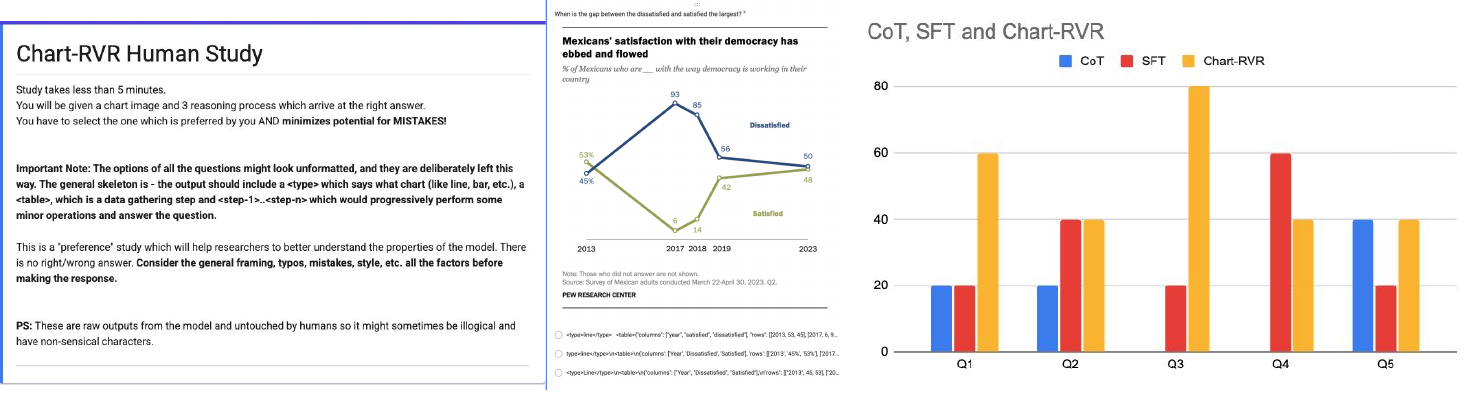}
    \caption{Layout and results of the human study. (LEFT) Instructions to the volunteers, (MIDDLE) Sample Question, and (RIGHT) Percentage of people preferring each of the 3 approaches.}
    \label{fig:human-study}
\end{figure*}

Finally, to ascertain the interpretability of the rationales generated by our approach as compared to SFT and CoT, we conduct a human study with 5 graduate-level volunteers. Our study has 5 data samples from the ChartQA and EvoChart datasets, which are \textbf{correctly} classified by all 3 approaches - CoT, SFT, and Chart-RVR, and the traces output by these approaches. We do not disclose which method produces what trace, and also shuffle the options. We show the instructions for the study, a sample question, and the final results in Figure~\ref{fig:human-study}. As can be observed, most people prefer Chart-RVR responses for 4/5 questions as compared to SFT and CoT, implying a clear preference. A more complete human study is reported in \url{https://arxiv.org/abs/2510.10973}.

\end{document}